\documentclass{IEEEtran}
\usepackage{amsthm}
\usepackage{amsmath}
\usepackage{amssymb}
\usepackage{graphicx}
\usepackage[unicode=true]{hyperref}
\usepackage{epstopdf}
\usepackage{algorithm}
\usepackage{algorithmic,color}

\newtheorem{remark}{Remark}

\algsetup{indent=2em}

\begin{document}

\title{Non-parametric Image Registration of Airborne LiDAR, Hyperspectral and 
Photographic \\ Imagery of Forests}

\author{
\IEEEauthorblockN {
  Juheon Lee$^{a,b}$,      
    Xiaohao Cai$^{a,b}$,         
    Carola-Bibiane Sch{\"o}nlieb$^{a}$,
    and  David Coomes$^{b}$
  }
  
\IEEEauthorblockA {
  $^{a}$Image Analysis Group, Department of Applied 
 Mathematics and Theoretical Physics (DAMTP),\\
 University of Cambridge, CB3 0WA.\\
  $^{b}$Forest Ecology and Conservation Group, Department of Plant Sciences,\\  
 University of Cambridge, CB2 3EA.
 }
}
\maketitle
\begin{abstract} 
There is much current interest in using multi-sensor airborne remote sensing to monitor the 
structure and biodiversity of forests. This paper addresses the application of non-parametric 
image registration techniques to precisely align  images obtained from multimodal imaging, 
which is critical for the successful identification of individual trees using object recognition 
approaches. Non-parametric image registration, in particular the technique of optimizing 
one objective function containing data fidelity and regularization terms, provides flexible 
algorithms for image registration.   Using a survey of woodlands in southern Spain as an example,  
we show that  non-parametric image registration can be successful  at fusing datasets  
when there is little prior knowledge about how the datasets are interrelated 
(i.e. in the absence of  ground control points). The validity of non-parametric registration 
methods in airborne remote sensing is demonstrated by a series of experiments. 
Precise data fusion is a prerequisite to accurate recognition of  objects within airborne imagery,  
so non-parametric image registration could make a valuable contribution to the analysis pipeline.  
\end{abstract}

\section{Introduction}\label{sec:introduction}

Airborne multimodal (multi-sensor) imaging is increasingly used to examine vegetation 
properties  \cite{bentoutou2005automatic,brunner2010earthquake}.   
The advantage of using multiple sensors is that each detects a different feature 
of the vegetation, so that collectively they provide a detailed understanding of the ecological 
processes  \cite{asner2007carnegie,dalponte2008fusion,huang2009estimating}. 
Specifically, Light Detection And Ranging (LiDAR) devices produce detailed point clouds of 
where laser pulses have been backscattered from surfaces, giving information 
on vegetation structure \cite{lefsky2002lidar,lim2003lidar};  
hyperspectral sensors measure reflectances within narrow wavebands, providing spectrally 
detailed information about the optical properties of targets \cite{asner2007carnegie,asner2008pnas}; 
while  aerial photographs  provide high spatial-resolution imagery  within three 
colour bands \cite{hudak1998textural,nakashizuka1995forest}. 
Using a combination of these sensors, individual trees in tropical rain forests can be 
mapped, enabling invasive species to be monitored \cite{asner2008pnas,asner2008invasive}, 
carbon storage to be assessed \cite{asner2009tropical} and leaf physiological processes 
to be inferred \cite{asner2011canopy,asner2007carnegie}. Accurate alignment of images 
is critical for the successful identification of individual trees using object recognition 
approaches \cite{asner2007carnegie,asner2008invasive,asner2008pnas}.  
However, images taken from different sensors or angles have relative rotation, 
translation or scale mismatches, and rugged terrain can cause complex 
displacement between images \cite{zitova2003image,le2011image}. As a result, 
aligning images is challenging. 

Alignment of remotely sensed images (known as image registration) is currently 
conducted with feature-based methods
\cite{brown1992survey,le2011image,li1995contour, yang2009remote,bentoutou2005automatic,lowe2004distinctive,li2009robust, 
goncalves2011automatic,wahed2013automatic,fischler1981random, kim2003automatic,goncalves2011automatic,gonccalves2011hairis,wong2007arrsi},
intensity-based methods \cite{zitova2003image,chen2003mutual,chen2003performance,
suri2010mutual,parmehr2014automatic,haber2006intensity,kroon2009mri,fonseca1996registration,hong2008wavelet} or a combination of the two  \cite{le2011image, liang2014automatic,ye2014local}, but all these approaches have their drawbacks.  
Image registration involves transforming a template image $T$ so that it aligns with 
a reference image $R$.   Feature-based methods rely on identifying common 
features in $R$ and $T$, for example ground control points (GCPs), patches or edges located 
in the images. These features are used to calculate 
transformation parameters, such that the location of the features in the 
transformed $T$ image are identical to those in $R$. Feature information can 
be obtained using  manual selection \cite{brown1992survey,le2011image}, 
edge detection \cite{li1995contour,yang2009remote,bentoutou2005automatic}, 
scale invariant feature transformation 
\cite{lowe2004distinctive,li2009robust,goncalves2011automatic,wahed2013automatic,ye2014local}, 
random sample consensus \cite{fischler1981random,kim2003automatic}, 
feature segmentation \cite{goncalves2011automatic,gonccalves2011hairis} or 
a phase congruency method \cite{wong2007arrsi}.  Feature-based methods 
can be very effective, but their performance relies on image quality and it can 
be difficult to locate corresponding features between images when datasets have 
different spatial resolutions or optical properties \cite{chen2003mutual,chen2003performance}. 
Furthermore, in the case of multimodal imaging, 
some features in the reference image may not be present in the template image, or vice versa. 
Intensity-based methods involve maximising the similarity in intensity 
values between the transformed template image $T_{reg}$ and $R$
\cite{zitova2003image,chen2003mutual,chen2003performance,suri2010mutual,parmehr2014automatic}. 
In theory this approach is fully automatic,  but in practice it  is often  mathematically ill-posed, 
in the sense that the registration solution might not be unique and a small change within 
the data might result in large variation in  registration results  \cite{hadamard1902problemes}. 
In addition, image modality affects the similarity between images significantly,  
therefore the choice of  similarity measure for the intensity-based methods is very 
important \cite{cole2003multiresolution,chen2003performance,suri2010mutual,parmehr2014automatic,
inglada2004possibility,bunting2008area,brunner2010earthquake,
liang2014automatic,ye2014local}. In general, it is extremely difficult to 
co-register multi-modal images if the images are not preprocessed 
(i.e. orthorectified or georeferenced). In this case, current intensity-based algorithms are likely to 
fail as they usually assume that displacement is neither complex nor large 
\cite{inglada2004possibility,bunting2008area,brunner2010earthquake,
liang2014automatic,ye2014local}. 
Therefore, preprocessing methods must be applied  to align images precisely. 
But preprocessing methods requires GCPs (i.e. feature information) 
\cite{lin2007map,berni2009thermal,turner2012automated,verhoeven2012mapping} 
unless the flight navigation system is fully integrated with orientation (bore-sight) calibrated 
imaging sensors \cite{laliberte2010acquisition,bryson2010airborne,turner2014direct}. 
Although there are near-automatic ways to get GCPs, it is still difficult to extract and 
choose common features from multi-modal images. Generally, GCPs or similarity measures are 
used to calculate optimal transformation parameters in affine 
transformation \cite{B87} (preserves points, straight lines and planes), 
which has been widely used in both feature-based and intensity-based registration methods \cite{modersitzki2009fair,song2014novel}.

This paper develops the use of a non-parametric registration method based on 
variational formulation as an alternative to the well-established feature-based and 
intensity-based approaches \cite{le2011image}.   Non-parametric registration 
methods are already well-established in mathematical analysis, medical imaging communities 
and computer vision
\cite{amit1994nonlinear,maintz1998survey,modersitzki2003numerical,zitova2003image,
modersitzki2009fair,burger2013hyperelastic} but have yet to permeate far in the field 
of remote sensing.  Unlike parametric image registration, which uses a small set of 
parameters to transform $T$ (examples include affine transformations calculated by
intensity-based Normalised Cross Correlation \cite{fonseca1996registration,hong2008wavelet},
Mutual Information \cite{cole2003multiresolution,chen2003performance,suri2010mutual,parmehr2014automatic} and Normalised Gradient
Fields \cite{haber2006intensity,kroon2009mri}),  
non-parametric registration methods are based on a variational formulation within which 
a cost function is minimised. They have been developed to overcome the ill-posedness 
of established methods by considering not only the similarity between images but also 
the regularity of the transformation in the calculated cost function, so that they can deal with 
non-linearity effectively 
\cite{broit1981optimal,amit1994nonlinear,maintz1998survey,zitova2003image,
modersitzki2003numerical,modersitzki2009fair}.  
To the best of our knowledge, these  methods have never been applied to the registration 
of remote sensing imagery.
  
We will demonstrate how non-parametric registration can be used to register three types  
of airborne remote sensing data sampled over forests 
(i.e. LiDAR, hyperspectral and photographic imagery). 
The benefits of the non-parametric registration method are illustrated, focussing particularly 
on its strong performance regardless of modality or degree of preprocessing. 
The datasets used to exemplify the approach are introduced in Section \ref{sec:data}. 
Then in Section \ref{sec:method}, the mathematical concepts of the non-parametric image 
registration algorithm are introduced. The demonstration of the effectiveness of 
the approach is given in Sections \ref{sec:app} and \ref{sec:experiments}.  
Finally, Section \ref{sec:con} gives recommendations for future
work. 

\section{Data}\label{sec:data}
This section briefly addresses the methodologies and properties of the datasets 
used for registration in this paper. Acquisition of remote sensing datasets was 
conducted in three areas of the 
Los Alcornocales Natural Park, Spain (lat 36$^{\circ}$19$'$, long 5$^{\circ}$37$'$) 
on 10 April 2011, by Airborne Research and Survey Facility of the UK's Natural  
Environment Research Council (NERC-ARSF) and preprocessed by their Data Analysis Node. The airplane 
flew at a nominal height above ground of approximate 3000~m and was equipped 
with LiDAR and hyperspectral imagers, as well as a digital camera. 
LiDAR [Leica ALS 50-II] emits pulses of monochromatic 
laser light  (1064 nm)  to scan topographical and geometrical structures of 
the surface, creating three-dimensional point clouds representing the points at 
which pulses are backscattered off surfaces and returned to the aircraft. 
Each point has an associated intensity value, which correlates with the 
proportion of a pulse's energy which is returned to the sensor. 
However, the radiometric properties of LiDAR intensity are not completely 
known - LiDAR pulse intensity values are controlled by an automatic gain 
control (AGC) system during the acquisition process,  so the intensity of the 
return is a function of unknown varying pulse energy as well as the backscattering 
properties of the ground surface 
\cite{kaasalainen2009radiometric, korpela2010range, korpela2010tree}.
NERC-ARSF preprocessed these LiDAR data and georeferenced them to the 
Universal Transverse Mercator (UTM) projection with WGS-84 datum. 
The average LiDAR point density over the study site is 2 points per square metre (m$^2$). 
In order to compare LiDAR imagery with other datasets, LiDAR point clouds were 
projected onto a two-dimensional image plane by ignoring the height information for each LiDAR point. LiDAR intensity was calculated 
in 1 m pixels as the average of the all-return pulse intensities.

Hyperspectral imaging spectrometers measure solar energy reflected off the 
earth's surface within a swath of land. Hyperspectral data were gathered using the 
AISA Eagle and AISA Hawk sensors (Specim Ltd., Finland) with 255 and 256 spectral 
bands respectively covering 400--2500 nm wavelengths across 2300~m 
swath width with 3 m spatial resolution. The hyperspectral sensors record reflected 
energy in digital numbers, which were converted to spectral radiance 
($\mu{\rm W}{\rm cm}^{-2}{\rm sr}^{-1}{\rm nm}^{-1}$) and then provided to us. 
Before image registration, hyperspectral imagery was atmospherically corrected 
using ATCOR-4 (Rese Ltd., Switzerland), which converts radiance values to reflectances. 
An accurate navigation system integrated with boresight calibrated hyperspectral sensors 
provide geocoordinates of each pixel in the hyperspectral imagery, which meant that the 
hyperspectral images could be orthorectified by digital elevation models (DEM) 
from ASTER and LiDAR data and then georeferenced to the UTM 
projection with WGS-84 datum. The estimated georeferencing error of hyperspectral image 
is about $5-10$ m horizontally. However, it deteriorates at the edge of 
the field of view of the hyperspectral sensors.

Aerial photographs were acquired during the flight using a Leica RCD-105 
Digital Frame Camera. Each photograph has $7212 \times 5408$ pixels. 
Since the spatial resolution of aerial photographs is much higher than that of 
hyperspectral images, aerial photos can help to identify objects more accurately. 
However, aerial photographs were not integrated with the aircraft navigation system, 
so they were not orthorectified or georeferenced during pre-processing. 
Metadata associated with aerial photographs informs of the time, location and 
altitude of aircraft when each photo was taken. We assumed that the location was the 
centre of each image and that the spatial resolution of each pixel equaled to 0.3 m.
  
If the preprocessed data had been georeferenced to 1~m 
accuracy then it would have been completely straightforward to register images, 
but the fact that uncertainty in the spatial resolution of the hyperspectral image 
often exceeds 3 m means that image registration techniques need to be be 
applied in order to precisely align images. Registration of aerial photos onto 
hyperspectral images or LiDAR imagery is even more challenging because 
they were neither orthorectified nor georeferenced when delivered. 
This paper provides a robust and accurate  approach for registering all three datasets.

\section{Method}\label{sec:method}

This section will briefly describe the mathematical concept of image registration, 
and the particular registration method that we use for the registration of images in 
our dataset (see \cite{fischer2004unified} for further details). Let $R$ and $T$ 
be the given reference and template images, respectively, modelled as functions 
defined on a finite two-dimensional grid $\Omega$ and mapping a point $x$ 
on the grid to a real intensity value $R(x)$ and $T(x)$, 
respectively. 
 
\begin{remark}
Note that the resolutions of $R$ and $T$ do not necessarily have to be the same, 
that is they can have different sizes in vertical and horizontal directions. 
As such, the grid $\Omega$ refers to a spatial domain on which both $R$ and $T$ 
are defined rather than the particular resolution of the latter. 
\end{remark}

When registering the template $T$ with the reference image $R$ we find a suitable 
transformation which maps $T$ to $R$ such that the transformed version of $T$ 
is similar to $R$. This transformation maps the grid of $T$ to the grid of $R$. 
A generic variational method for finding this transformation is as a solution 
$\varphi:\Omega\rightarrow\Omega$ of
\begin{align}  \label{eq:varmodel} 
\min_\varphi\left\{\sum_{{x}\in\Omega} D[T(\varphi({x}))),R({x})] 
+ \alpha S(\varphi)\right\},
\end{align}
where $D$ is a similarity measure that quantifies the difference between the 
distorted template $T$ and reference image $R$, $S$ is a so-called regularisation 
term that imposes appropriate regularity on the transformation $\varphi$ and $\alpha$ 
is a positive parameter that balances the importance of the similarity measure against 
the regularisation term. Existence of solutions of \eqref{eq:varmodel}
for the registration problem are discussed, for example, 
in \cite{modersitzki2003numerical,modersitzki2009fair,fischer2003curvature} 
and the references therein.
In the particular case of non-parametric registration considered 
in this paper, the transformation function $\varphi$ can be expressed as the sum of 
identity and displacement $u$, that is
\begin{align}
\varphi: {x} \rightarrow {x} - {u(x)}. 
\end{align}
A standard choice for $D$ in \eqref{eq:varmodel} is
$$
D[T(\varphi({x}))),R({x})] = \frac{1}{2} [T({x}-{u})-R({x})]^2,
$$
which has the disadvantage of not being  contrast-invariant \cite{modersitzki2009fair}. 
This can be corrected by using gradient information rather than intensity information 
to measure similarity \cite{modersitzki2009fair}. 
In this paper we use a NGF similarity 
measure \cite{haber2006intensity,modersitzki2009fair}. 
Here, the normalized gradient $\frac{\triangledown I}{|\triangledown I|}$ of 
an image $I$ is used to measure similarity between $R$ and $T$. 
More precisely, the NGF measure is defined as
\begin{align} 
{\rm NGF}(I,\eta) = {\tt vec}\left(\frac{\nabla I}{\sqrt{|\nabla I|^2 + \eta^2}} \right)
\end{align}
where $\eta$  is edge parameter $(\eta > 0)$, and ${\tt vec}(\cdot)$ is the command of generating a vector by aligning the columns of the input. The edge parameter $\eta$ models the level of 
the noise present in $I$ such that image values below this parameter are ignored. 
Then NGF distance measure is defined as
\begin{align}  \label{eq:termfid} 
D^{\rm NGF}[T(\varphi({x})),R({x})] 
= 1- \left(\left({\rm NGF}(T,\eta)\right)^T {\rm NGF}(R,\eta)\right)^2,
\end{align}
which, if minimised, maximises the linear dependency (alignment) of the NGF 
of $T$ and $R$. A number of other similarity measures have been suggested 
for different types of image analysis, cf. \cite{modersitzki2009fair}. 

The regularisation term $S$ encodes the regularity that should be imposed on the 
transformation $\varphi$ to reduce the ill-posedness of the regstration problem. 
For an overview of different regularisation terms and their effect on the registration, see 
\cite{modersitzki2009fair,modersitzki2003numerical}. 
In what follows we use a curvature regularisation 
\cite{fischer2003curvature,haber2006intensity}, that is
\begin{align} \label{eq:termregu} 
S^{curv}(\varphi) = S^{curv}({u}) 
= \frac{1}{2}\sum_{{x}\in\Omega} |\bigtriangleup{u}({x})|^2.
\end{align}
This regularisation results in the registration accuracy being dependent on the 
smoothness of the displacement ${u}$ between $R$ and $T$ \cite{modersitzki2003numerical}. 
In particular, curvature regularisation penalises oscillations in ${u}$ 
since it can be regarded as an approximation of the curvature of 
${u}$ \cite{modersitzki2003numerical}. One advantage of 
curvature regularisation is that it does not require affine preregistration steps. 
Other regularisation techniques, such as fluid registration 
\cite{christensen1994deformable,bro1996fast}, are sensitive to affine 
linear displacement such that preregistration with affine linear transformation is 
required, see \cite{modersitzki2003numerical,fischer2003curvature,fischer2004unified}. 
With these choices for $D$ in \eqref{eq:termfid} and $S^{curv}$ in \eqref{eq:termregu},
this leads the registration method on the minimisation of the specific functional
\begin{align} \label{eq:registfunc} 
J({u}) = 
\sum_{{x}\in\Omega} D^{\rm NGF}[T(\varphi({x})),R({x})] 
+ \frac{\alpha}{2} \sum_{{x}\in\Omega} |\bigtriangleup  u(x)|^2. 
\end{align}

For the numerical minimisation of \eqref{eq:registfunc} we use the 
\emph{Image Registration software package} (FAIR)\footnote{MATLAB version of 
FAIR \url{http://www.siam.org/books/fa06/}}. There, the minimiser of \eqref{eq:registfunc} 
is computed iteratively via a semi-implicit scheme for the so-called Euler-Lagrange 
equation for \eqref{eq:registfunc}. The latter is the equation that arises as the spatially 
discrete version of the G\^{a}teaux derivative of the continuous functional $J$, 
which reads \cite{fischer2003curvature}
\begin{align}
\label{eq:eulerlagrange} {f(x,u(x))} + \alpha \bigtriangleup^2 {u(x)} 
= 0 \qquad \text{ for } {x} \in \Omega,
\end{align}
where ${f(x,u(x))}$  is the discretisation of the derivative of the distance measure 
$D$. In order to solve equation \eqref{eq:eulerlagrange} a semi-implicit iterative 
scheme is used which introduces an artificial time step $\Delta t$ and computes the fixed 
point of the equation \cite{modersitzki2003numerical,fischer2003curvature,modersitzki2009fair}
\begin{align} 
{u}^{k+1}({x}) - \Delta t ~\alpha \Delta^2 {u}^{k+1}({x})
 ={u}^{k}({x})  + \Delta t~{f}({x},{u}^{k}({x})), 
 \end{align}
where ${u}^{k}({x})$ denotes the $k$-th iterate of the scheme.
Further details regarding discretisation and numerical optimisation are provided 
in \cite{modersitzki2009fair}. Since remote sensing datasets contain large-scale 
surface information, it is computationally expensive to conduct entire image registration 
steps at the original resolution \cite{modersitzki2009fair}. 
FAIR provides a multilevel image-registration scheme, producing a series of images 
varying in resolution, such that registration results from a coarser image can be used 
to initialise the registration on finer resolutions of the images. 
The multilevel scheme reduces the expensive computation and
 the chance of being trapped in local minima during the 
iterative search as images are much smoother in coarse resolution, cf. 
\cite{haber2006multilevel,modersitzki2009fair}.

\section{Application of the Registration Approach to the Airborne Remote Sensing Dataset}
\label{sec:app}

The first step of the process was to match the geographical boundaries of all datasets to each other, reducing the number of features present in either $R$ or $T$, but not both. Since both hyperspectral and LiDAR intensity images contain geo-coordinates, geographical boundary matching of them is straightforward. But the aerial photographs were neither georeferenced nor orthorectified and matching 
the boundary between aerial photographs and other datasets was therefore challenging. 
For the latter we used the geocoordinate at which each photo was taken as the centre 
of each aerial photograph. Then the geographic boundary of each aerial photo 
was roughly calculated by counting the approximate number of pixels of an aerial 
photograph and adding $300$ m in $x$ and $y$ directions to 
compensate the errors caused by rough approximation. Hence the size of each 
aerial photograph image was assumed to be 
$L_{r_x} \times L_{r_y}$ m$^2$, i.e. 
\[
L_{r_x} = 0.3 L_{t_x} +300, \quad L_{r_y} = 0.3 L_{t_y} +300,
\]
where $L_{t_x}$ and $L_{t_y}$ are the number of pixels of aerial photographs in 
$x$ and $y$ directions, and approximately equal to $7000$ and $5000$ 
respectively for the data tested in this paper. 

LiDAR is the most accurately georeferenced of the three datasets from the 
airborne sensors (having $0.1$ to $0.15$ m 
horizontal error and about 0.2 m vertical error). 
Therefore it is used as the reference image onto which the hyperspectral 
template image is aligned. LiDAR intensity data and the mean intensity of 
RGB bands (640, 549 and 460 nm) of the hyperspectral images were used. 
Although it would seem natural to use the band at 1065 nm wavelength 
of the hyperspectral imagery -- which corresponds to the LiDAR intensity 
wavelength -- this band suffers from low signal-to-noise ratio. 

Non-parametric image registration with a variational formulation finds the 
optimised location for each pixel, which maximises similarity between two images. 
This can be achieved by numerical optimisation methods, the choice of which 
can influence the performance of image registration. The FAIR toolbox provides 
three different second-order optimisation schemes: Gauss-Newton, l-BFGS and 
Trust region, all of which were explored (see experimental results). 
Non-parametric registration yielded optimised spatial coordinates of each pixel, 
which were used for the transformation of original hyperspectral images. 
During the transformation, the hyperspectral images were interpolated and 
resampled by nearest neighbour interpolation. Interpolation estimates were 
chosen from existing values, thus minimising interpolation artefacts. 
This is important because hyperspectral imagery should preserve physically 
meaningful values.

Choosing optimal parameters in \eqref{eq:registfunc} is the most important step of the 
registration process, but these are difficult to find automatically (although see 
\cite{ascher2006effective,haber2006intensity} for examples of automatic 
edge parameter $\eta$  selection once noise level and image volume are known).
We used a trial-and-error approach  to find $\eta$  and smoothness parameter $\alpha$, 
which was time consuming. Fortunately, tuning of parameters 
for each registration of remote sensing images is not normally required - a single 
calibration for template and reference images taken by two different sensors 
was enough to obtain reasonable results in most cases. 
For the registration of a hyperspectral image onto a LiDAR intensity image 
the optimal values of $\alpha$ and $\eta$ were found to be $5000$ and $0.1$,
respectively. 

The aerial photo was aligned first with the hyperspectral image and then from there 
registered onto LiDAR. We found this circuituous route necessary because the swath 
width of LiDAR ($800$ m) is much smaller than scene width of the aerial 
camera ($2400$ m). The registration of the aerial photographs 
onto the hyperspectral images is challenging because aerial photographs are 
distorted by various effects, among them topography, lense distortion or the viewing 
angle. As we regarded the location where each aerial photo was taken as the centre of 
the image, corresponding hyperspectral images of size $L_{r_x} \times L_{r_y}$ m$^2$ 
were extracted from the hyperspectral imagery and used as the reference image. 
Curvature registration with NGF distance measure \eqref{eq:registfunc} was employed 
to register aerial photographs onto hyperspectral  images. Regularisation parameter 
was set to $\alpha=1.5\times 10^{5}$ and the edge parameter $\eta =0.03$. 
RGB bands of hyperspectral images and RGB aerial photos were both transformed 
to grey intensity images before registering them to each other to increase the processing 
speed and the robustness of the registration. The results of the registration 
suggest that the method can handle both orthorectification and registration 
(see examples in Section \ref{sec:experiments}). After the registration of 
the aerial photographs onto the hyperspectral imagery, a mosaic of the aerial photos was 
created which was then aligned with the LiDAR data in an additional registration step. 
This last registration step was aided by the fact that hyperspectral and LiDAR imagery 
had already been aligned.

Numerical experiments were conducted, to compare our non-parametric 
approach (NP) \eqref{eq:registfunc} with  well-known parametric registration 
methods based on different distance measures (i.e. NCC \cite{fonseca1996registration,hong2008wavelet}, 
MI \cite{cole2003multiresolution,chen2003performance,suri2010mutual,parmehr2014automatic} 
and NGF \cite{haber2006intensity,kroon2009mri}).

\section{Experimental Results}\label{sec:experiments}

l-BFGS was found to be the fastest and most accurate of the second-order optimisation 
approaches available in the FAIR toolbox, and was used in all numerical experiments. 

\begin{figure*}[!htb]
\begin{center}
\begin{tabular}{ccc}
\includegraphics[width=40mm, height=40mm]{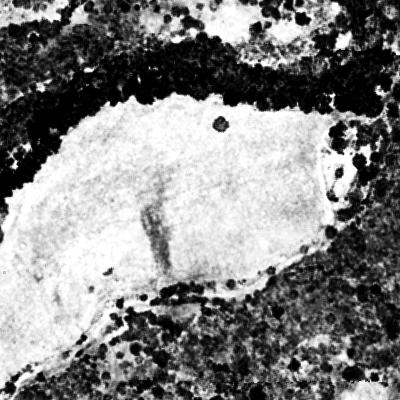} &
\includegraphics[width=40mm, height=40mm]{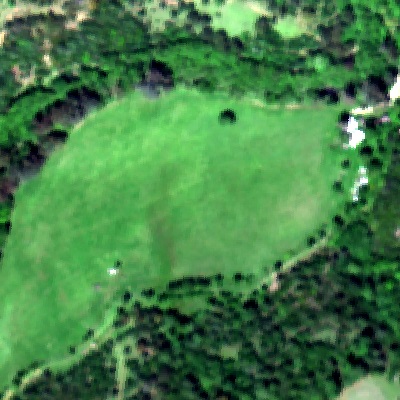} &
\includegraphics[width=40mm, height=40mm]{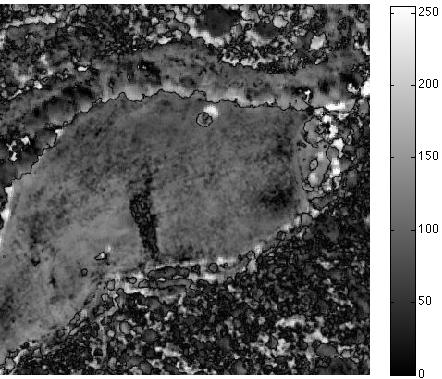} 
\put(-60,79){{\color{yellow} \circle{20,10}}} 
\put(-30,75){{\color{yellow} \circle{20,10}}}\\
(a) LiDAR: $R$ & (b) Hyperspectral: $T$ & (c) $\vert T-R \vert $ (73.7)
\end{tabular}
\begin{tabular}{cccc}
\includegraphics[width=40mm, height=40mm]{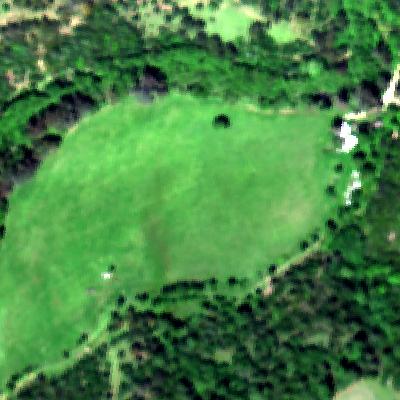}  &
\includegraphics[width=40mm, height=40mm]{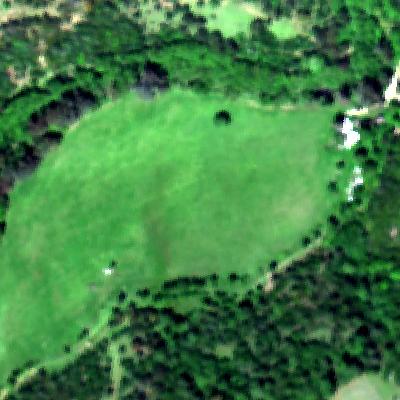} &
\includegraphics[width=40mm, height=40mm]{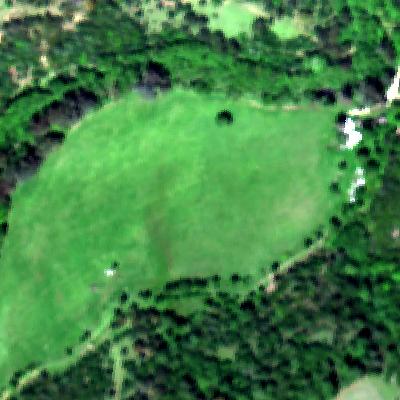}  &
\includegraphics[width=40mm, height=40mm]{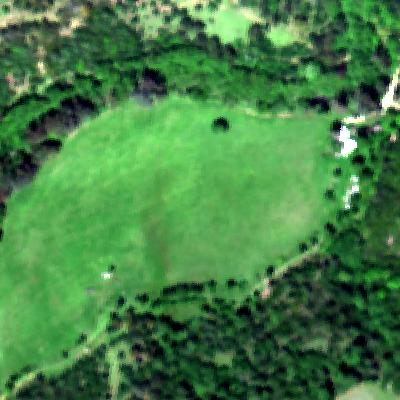}\\
 (d) NCC: $T^{\rm NCC}_{reg}$ & (e) MI: $T^{\rm MI}_{reg}$ &  
 (f) NGF: $T^{\rm NGF}_{reg}$ & (g) NP \eqref{eq:registfunc}: $T^{\rm NP}_{reg}$\\ 
\includegraphics[width=40mm, height=40mm]{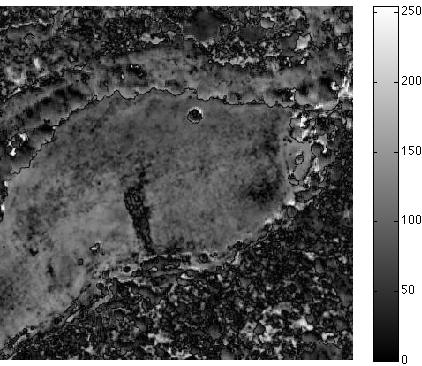} 
\put(-62,79){{\color{yellow} \circle{20,10}}} 
\put(-30,75){{\color{yellow} \circle{20,10}}}&
\includegraphics[width=40mm, height=40mm]{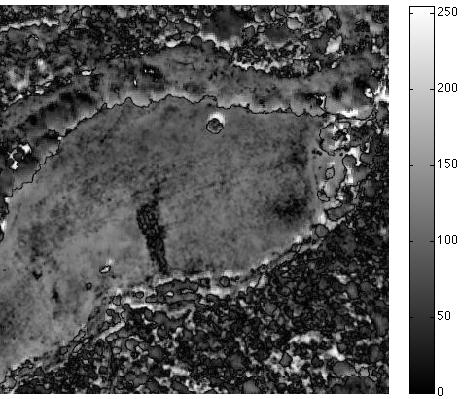} 
\put(-62,79){{\color{yellow} \circle{20,10}}} 
\put(-30,75){{\color{yellow} \circle{20,10}}}&
\includegraphics[width=40mm, height=40mm]{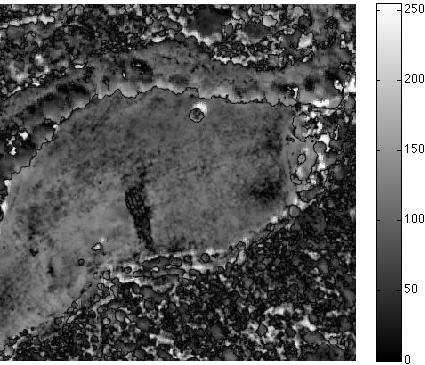}   
\put(-62,79){{\color{yellow} \circle{20,10}}} 
\put(-30,75){{\color{yellow} \circle{20,10}}}&
\includegraphics[width=40mm, height=40mm]{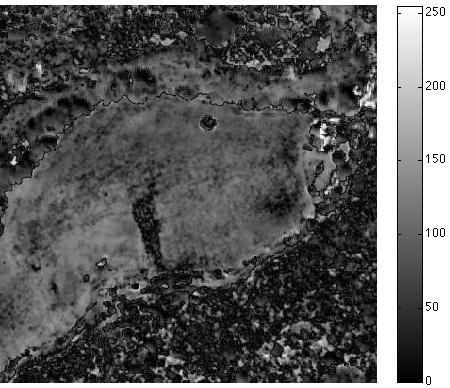} 
\put(-62,79){{\color{yellow} \circle{20,10}}} 
\put(-30,75){{\color{yellow} \circle{20,10}}}\\
 (h) $\vert T^{\rm NCC}_{reg}-R \vert$ ({66.9}) & (i)  $\vert T^{\rm MI}_{reg}-R \vert$ (73.1)
 & (j) $\vert T^{\rm NGF}_{reg}-R \vert$ (72.6) & (k) $\vert T^{\rm NP}_{reg}-R \vert$ (65.7)
\end{tabular}
\end{center}
\caption{Image registration of a hyperspectral image 
onto a LiDAR intensity image of a Spanish woodland, surveyed from an aircraft 
(scale $400 \times 400$ m$^2$). The first row shows (a) a LiDAR intensity reference image ($R$); 
(b) a hyperspectral template image ($T$); (c) a map highlighting  difference between these images 
(i.e. the complement of intensity differences  $\vert T-R \vert$), which would be entirely white if 
the match was perfect. The second row of panels 
show the aerial photograph template image after it has been registered using parametric 
methods (d) NCC, (e) MI and (f) NGF and the NP approach (g), the results of which are denoted by 
$T^{\rm NCC}_{reg}$, $T^{\rm MI}_{reg}$, $T^{\rm NGF}_{reg}$, respectively, 
and our non-parametric approach as $T^{\rm NP}_{reg}$. The final row of 
maps (h)--(k) highlight the absolute values of the differences between the registered hyperspectral 
images and the LiDAR reference image; the average intensity difference within the image is given in parentheses, indicating that NCC and NP approaches are similarly good whilst MI and NGF approaches are only slightly better than using the original template to calculate intensity differences (i.e. ${\tt mean}(\vert T-R \vert) = 73.7$). 
Yellow circles highlight regions of the images where differences among registration methods are seen.
}\label{example1}
\end{figure*}

The first case we consider is image registration of hyperspectral imagery onto 
LiDAR (Figure \ref{example1}). As both datasets were georeferenced by the data provider,  
only small distortions were present (up to $10$ m) as a result of DEM or navigation 
inconsistencies \cite{inglada2004possibility,bunting2008area,brunner2010earthquake,
liang2014automatic,ye2014local}.
Figure \ref{example1} (a) and (b) show the LiDAR intensity reference image ($R$) and 
hyperspectral template image ($T$), respectively, where the colour intensity of $T$ 
is the composite intensity of the RGB bands of the hyperspectral image. 
(c) is the complement of an intensity difference map between the LiDAR intensity reference 
image and the hyperspectral image. Figure \ref{example1} (g) shows the registration 
result using the non-parametric approach and (k) shows the intensity difference map
(the absolute values of the difference between the registered image and the 
reference image). The white to black pixel values represent small to large differences. 
From Figure \ref{example1} (h)--(k), in particular the parts inside 
the circles marked on the figures and the average value of the intensity of all pixels in 
each intensity difference map, 
we see that the results of the NCC and NP methods are better than the results of 
the MI and NGF methods. In this example, the NCC method 
performed as well as the NP method, because both the hyperspectral and LiDAR images 
were approximately georeferenced before the registration was applied, so finding a local 
minimum was enough to get reasonable outcomes \cite{brunner2010earthquake}.

\begin{figure*}[!htb]
\begin{center}
\begin{tabular}{ccc}
\includegraphics[width=40mm, height=55mm]{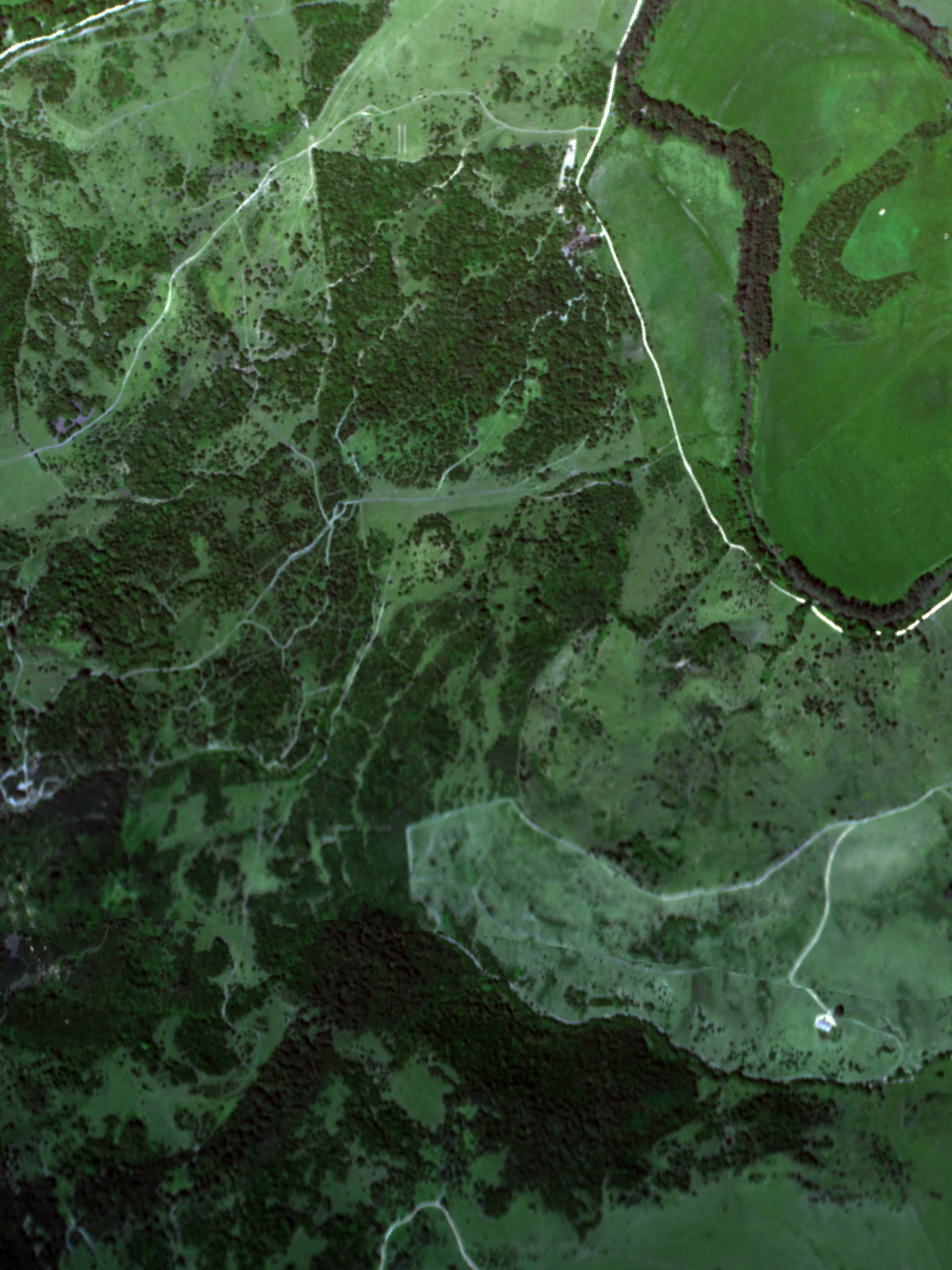}  &
\includegraphics[width=40mm, height=55mm]{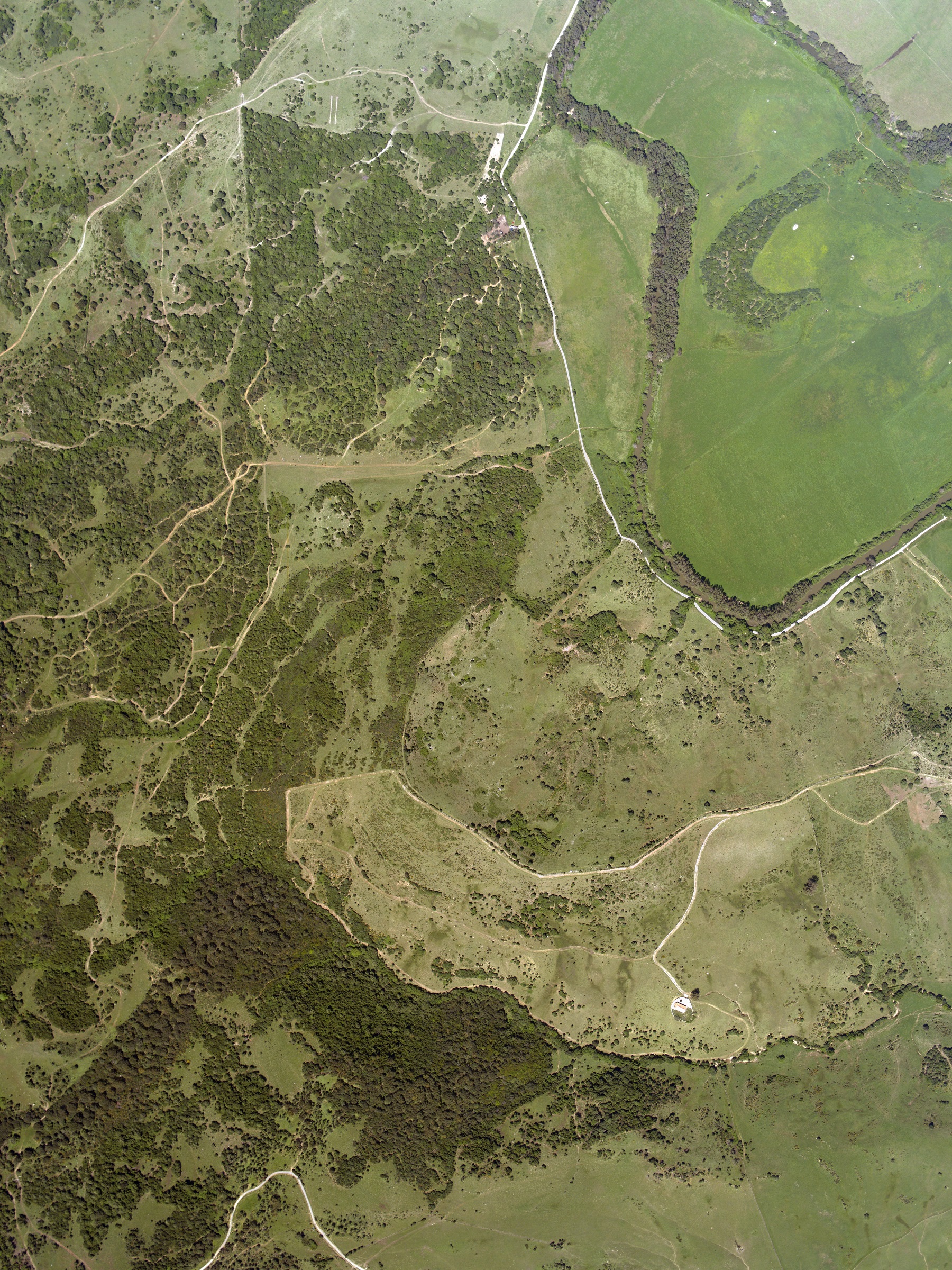}  &
\includegraphics[width=40mm, height=55mm]{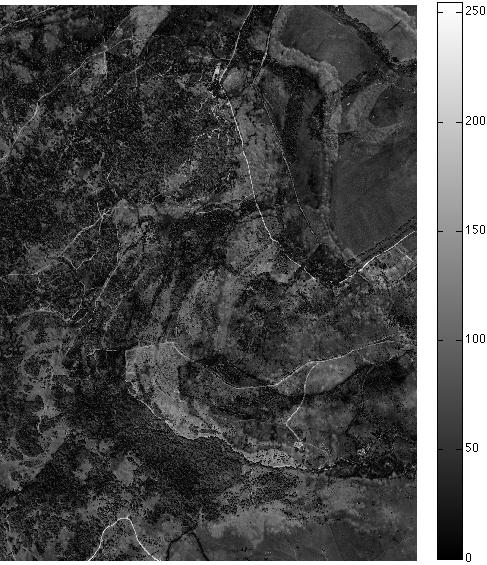} 
\put(-30,120){{\color{yellow} \circle{25,10}}} 
\put(-75,120){{\color{yellow} \circle{25,10}}} 
\put(-30,80){{\color{yellow} \circle{25,10}}} \\
(a) Hyperspectral: $R$  &  (b) Aerial photograph: $T$ & (c) $\vert T-R \vert$ (52.9393) 
\end{tabular}
\begin{tabular}{cccc}
\includegraphics[width=40mm, height=55mm]{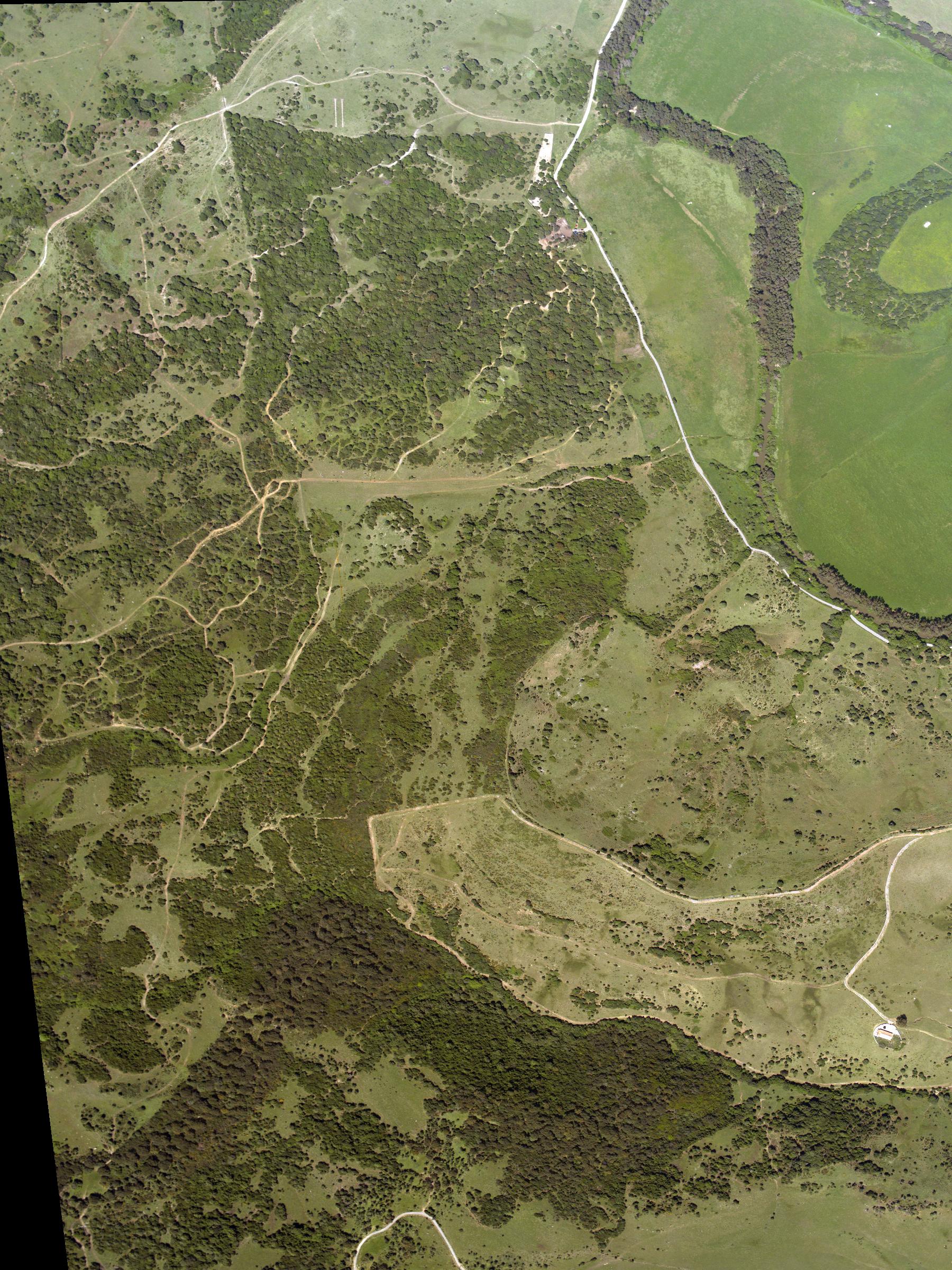}  &
\includegraphics[width=40mm, height=55mm]{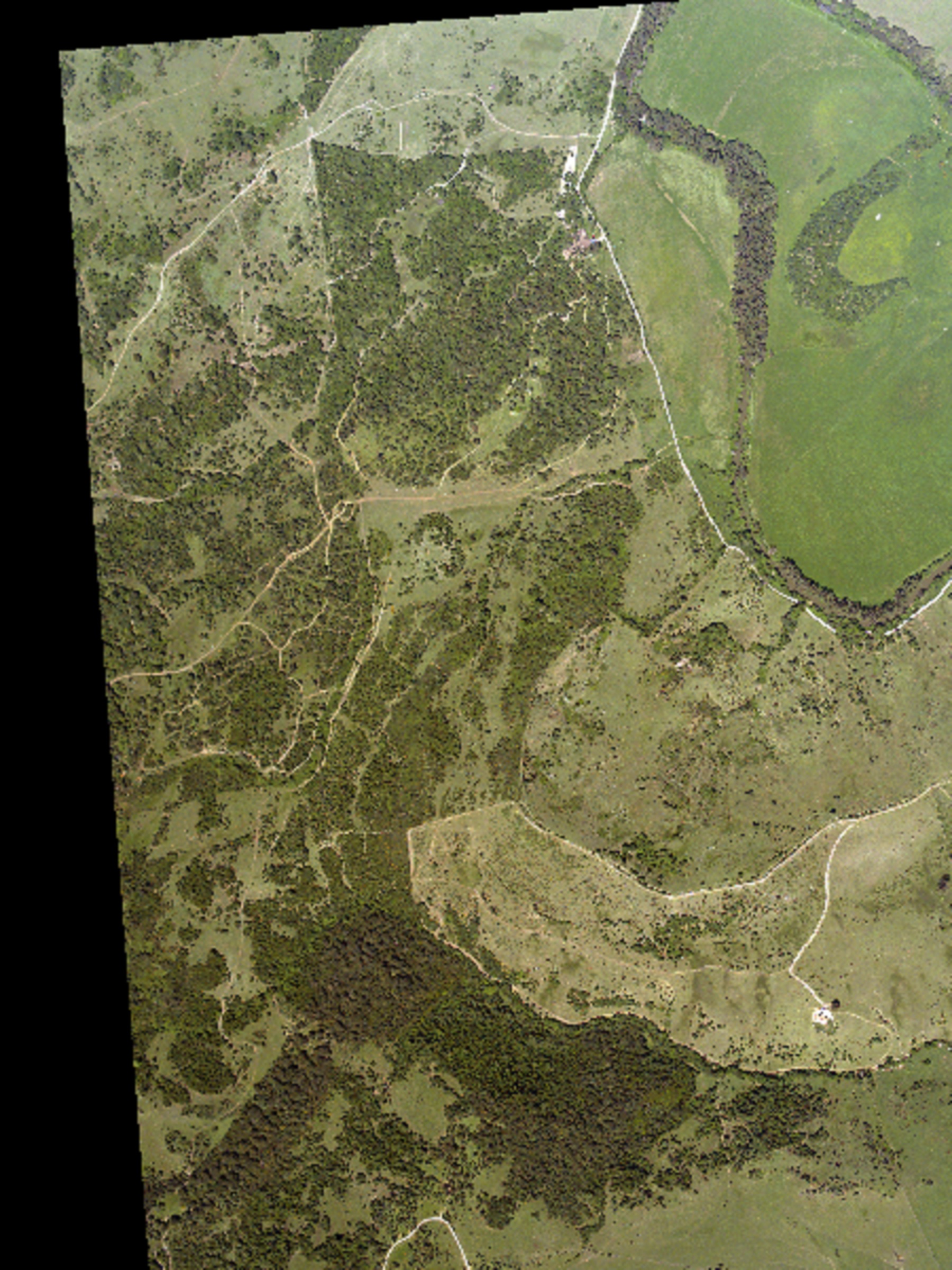} &
\includegraphics[width=40mm, height=55mm]{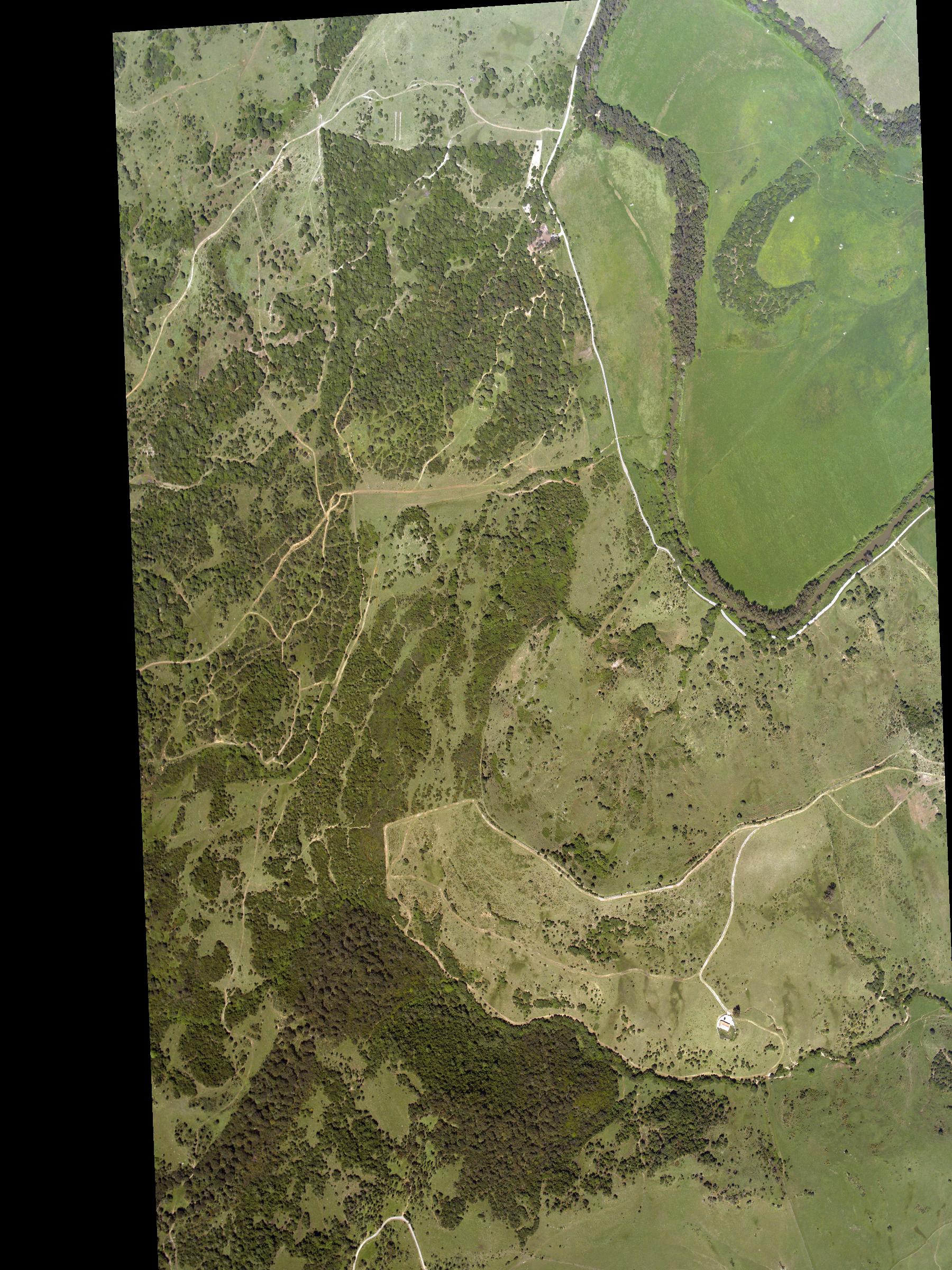}  &
\includegraphics[width=40mm, height=55mm]{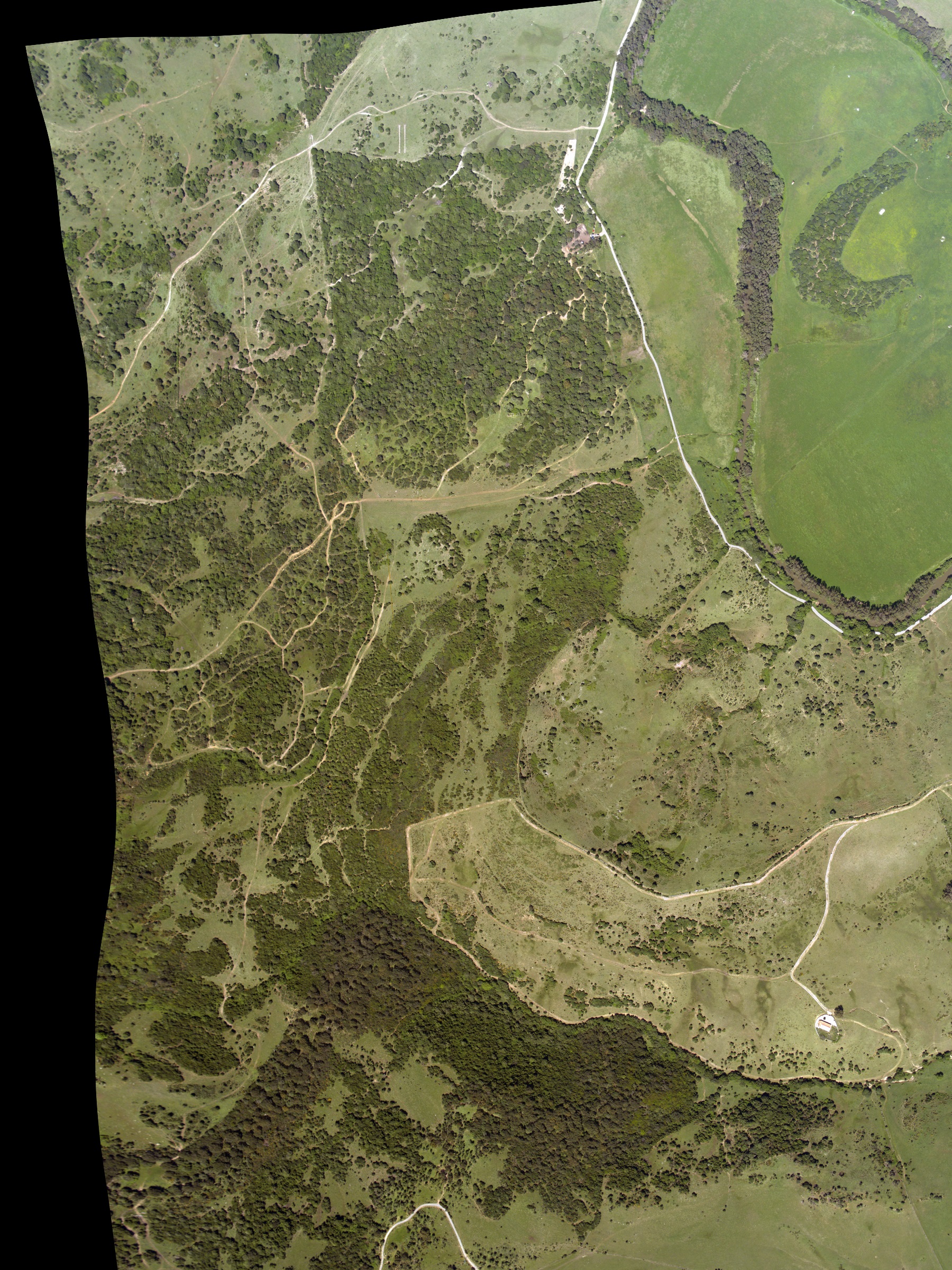} \\
 (d) NCC: $T^{\rm NCC}_{reg}$ & (e) MI: $T^{\rm MI}_{reg}$ &  
 (f) NGF: $T^{\rm NGF}_{reg}$ & (g) NP \eqref{eq:registfunc}: $T^{\rm NP}_{reg}$\\ 
\includegraphics[width=40mm, height=55mm]{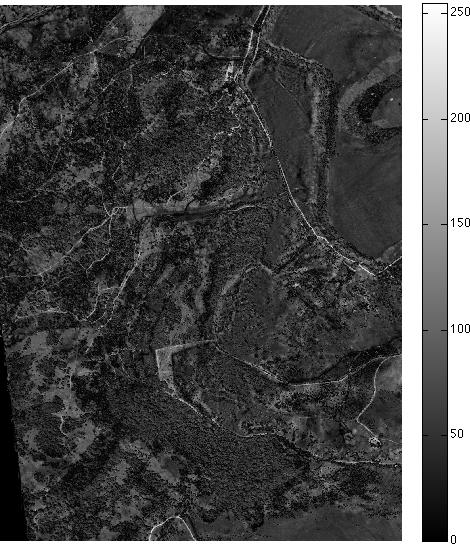} 
\put(-30,120){{\color{yellow} \circle{25,10}}} 
\put(-75,120){{\color{yellow} \circle{25,10}}} 
\put(-30,80){{\color{yellow} \circle{25,10}}}  &
\includegraphics[width=40mm, height=55mm]{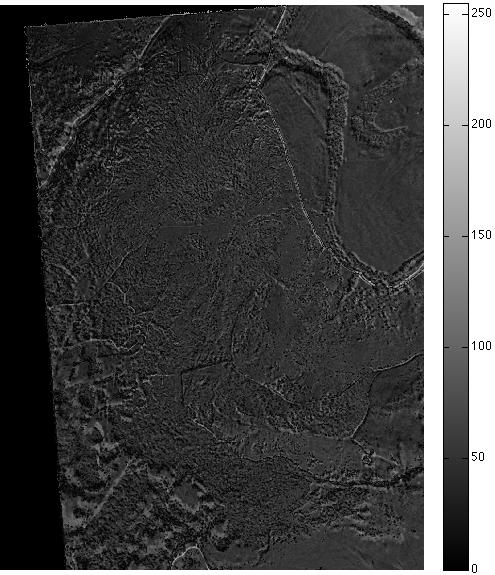} 
\put(-30,120){{\color{yellow} \circle{25,10}}} 
\put(-75,120){{\color{yellow} \circle{25,10}}} 
\put(-30,80){{\color{yellow} \circle{25,10}}}  &
\includegraphics[width=40mm, height=55mm]{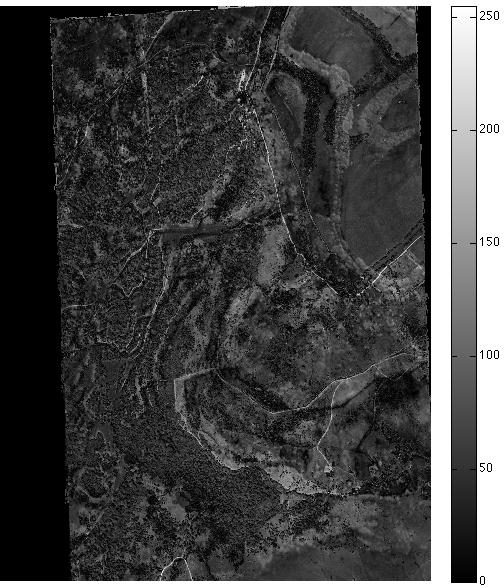}  
\put(-30,120){{\color{yellow} \circle{25,10}}} 
\put(-75,120){{\color{yellow} \circle{25,10}}} 
\put(-30,80){{\color{yellow} \circle{25,10}}}  &
\includegraphics[width=40mm, height=55mm]{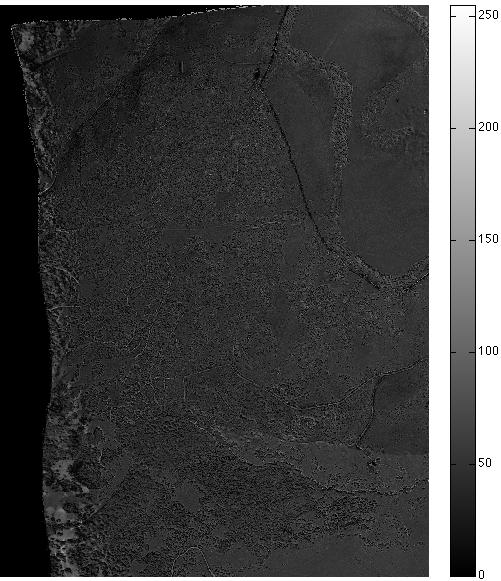} 
\put(-30,120){{\color{yellow} \circle{25,10}}} 
\put(-75,120){{\color{yellow} \circle{25,10}}} 
\put(-30,80){{\color{yellow} \circle{25,10}}}    \\
 (h) $\vert T^{\rm NCC}_{reg}-R \vert$ (49.0) & (i)  $\vert T^{\rm MI}_{reg}-R \vert$ (46.2)
 & (j) $\vert T^{\rm NGF}_{reg}-R \vert$ (50.7) & (k) $\vert T^{\rm NP}_{reg}-R \vert$ (45.6)
\end{tabular}
\end{center}
\caption{Image registration of an aerial photograph onto a hyperspectral image 
in the case of flat terrain (scale $2400 \times 1800$ m$^2$). The first row shows 
(a) a hyperspectral reference image ($R$); (b) an aerial photograph template image ($T$); 
(c) a map highlighting  differences between these images (i.e. the complement of intensity 
differences  $\vert T-R \vert$). The second row of panels show the aerial photograph 
template image after it has been registered using parametric methods (d) NCC, (e) MI and (f) 
NGF and (g) the NP approach. The final row shows maps highlighting the absolute values of the
differences between the registered aerial photograph images and the hyperspectral 
reference image; the average intensity difference within the image is given in parentheses. }\label{example-flat}
\end{figure*}

\begin{figure*}[!htb]
\begin{center}
\begin{tabular}{ccc}
\includegraphics[width=40mm, height=55mm]{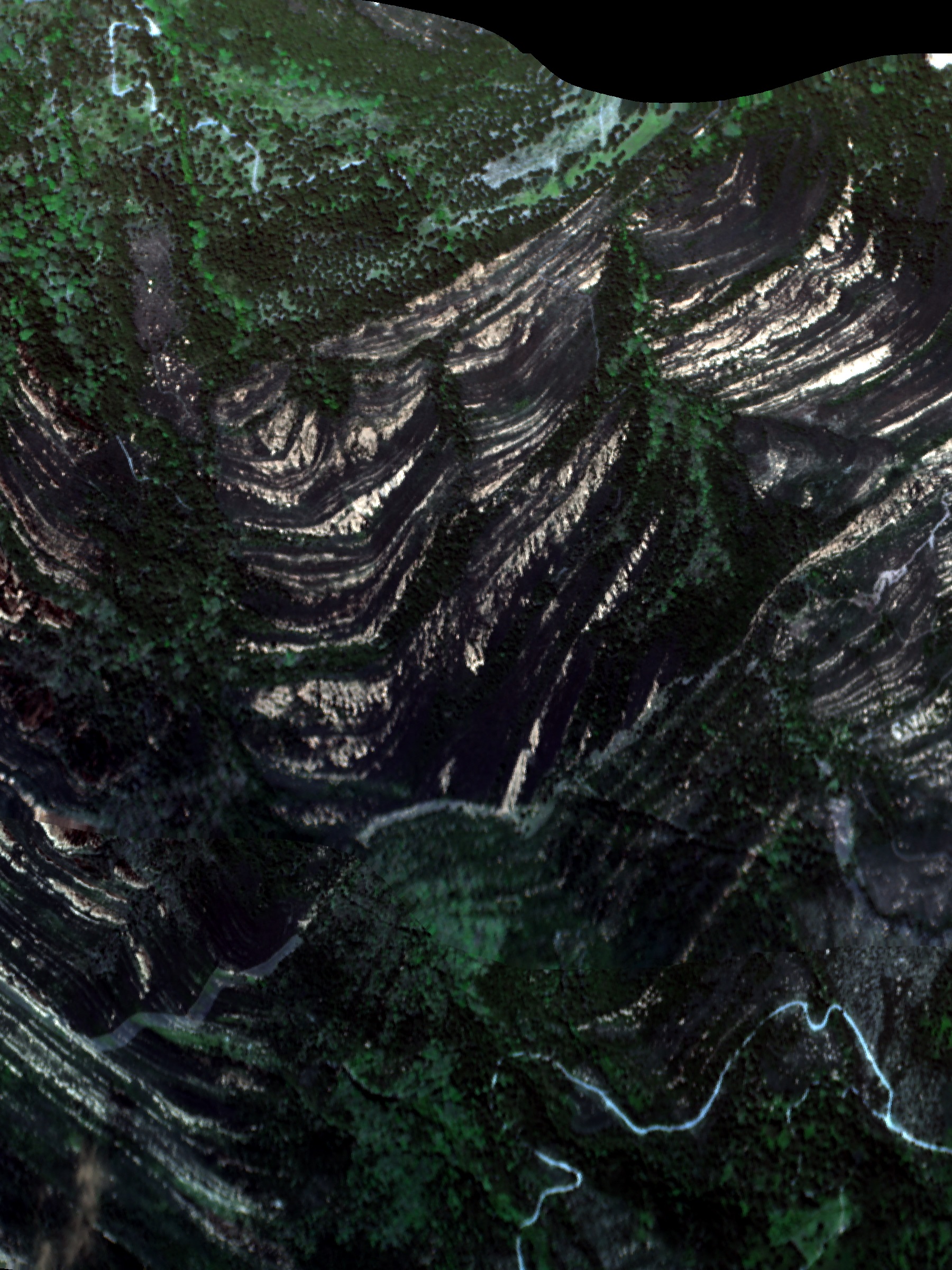}  &
\includegraphics[width=40mm, height=55mm]{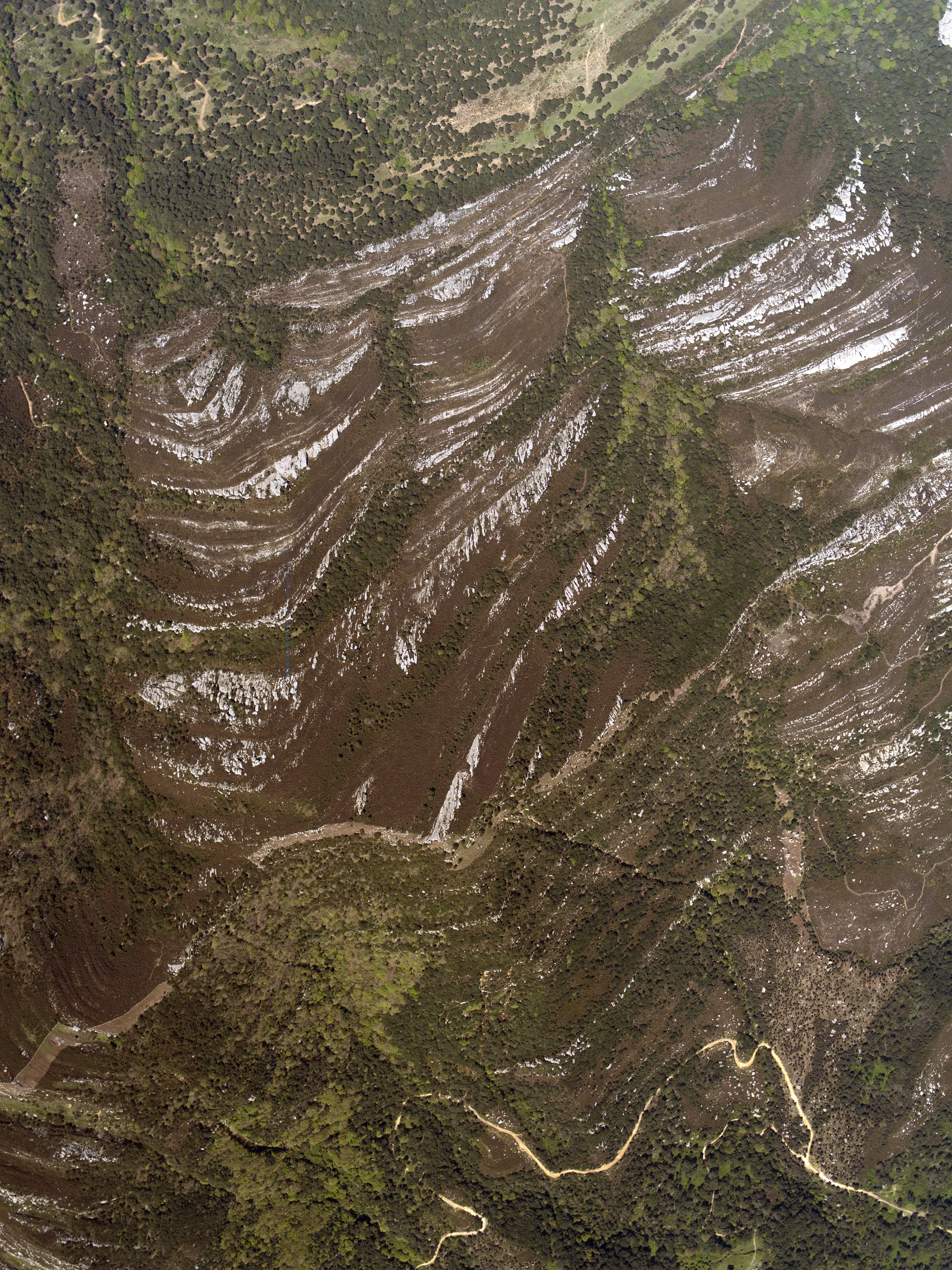}  &
\includegraphics[width=40mm, height=55mm]{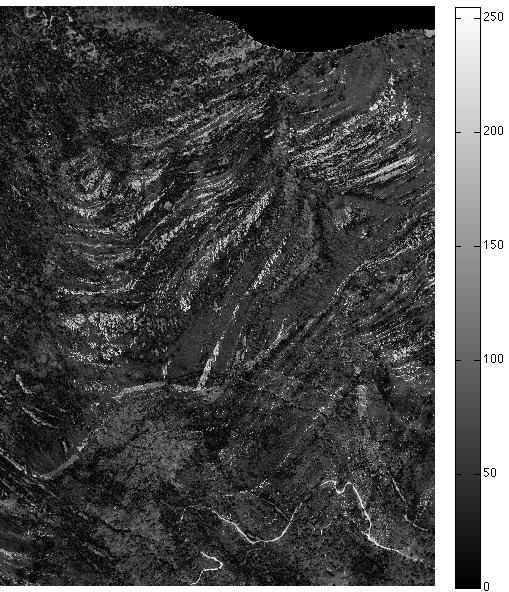}  
\put(-65,57){{\color{yellow} \circle{25,10}}} 
\put(-70,90){{\color{yellow} \circle{25,10}}} 
\put(-35,28){{\color{yellow} \circle{25,10}}}\\
(a) Hyperspectral: $R$  &  (b) Aerial photograph: $T$ & (c) $\vert T-R \vert$ (53.2706) 
\end{tabular}
\begin{tabular}{cccc}
\includegraphics[width=40mm, height=55mm]{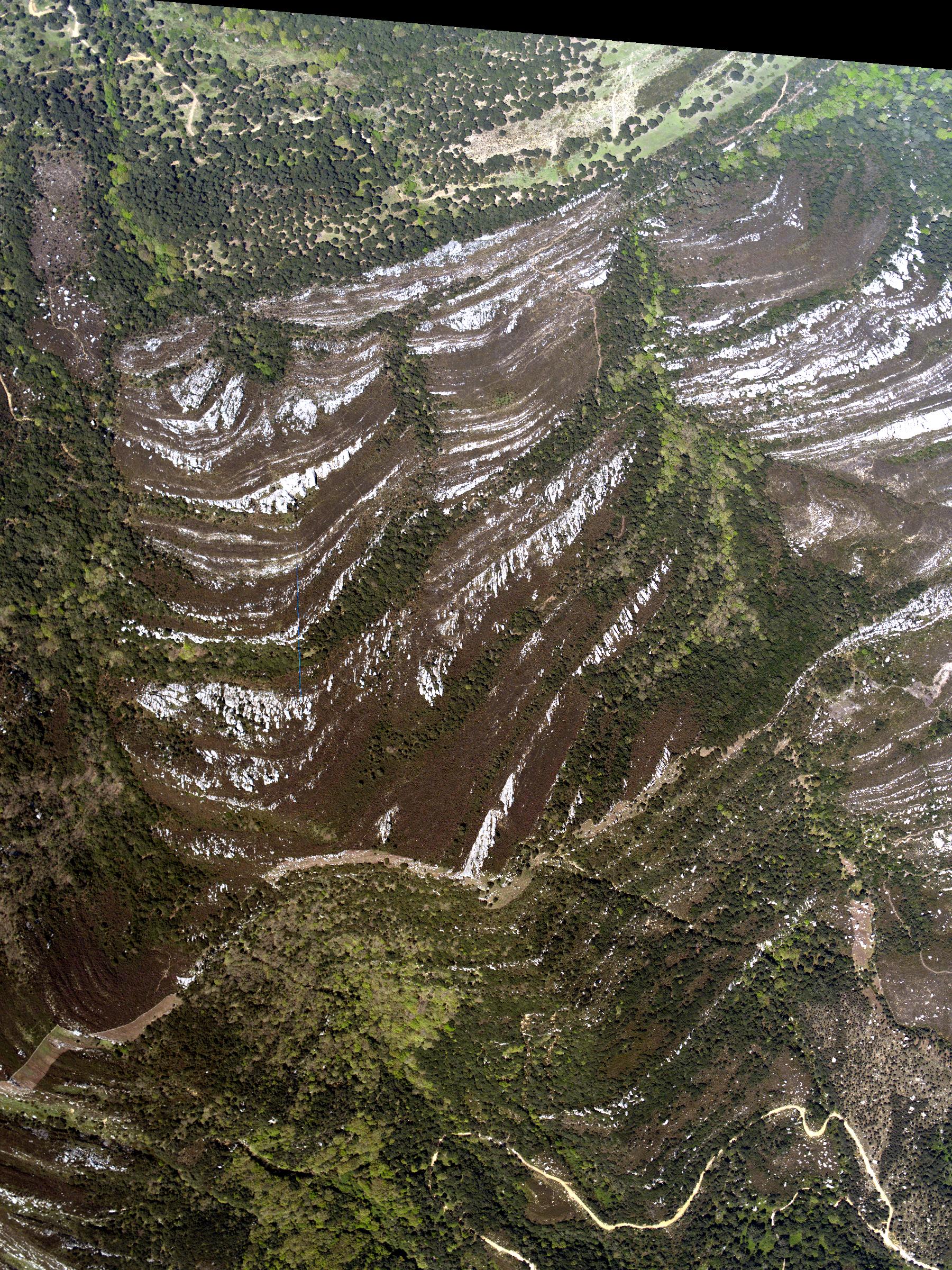}  &
\includegraphics[width=40mm, height=55mm]{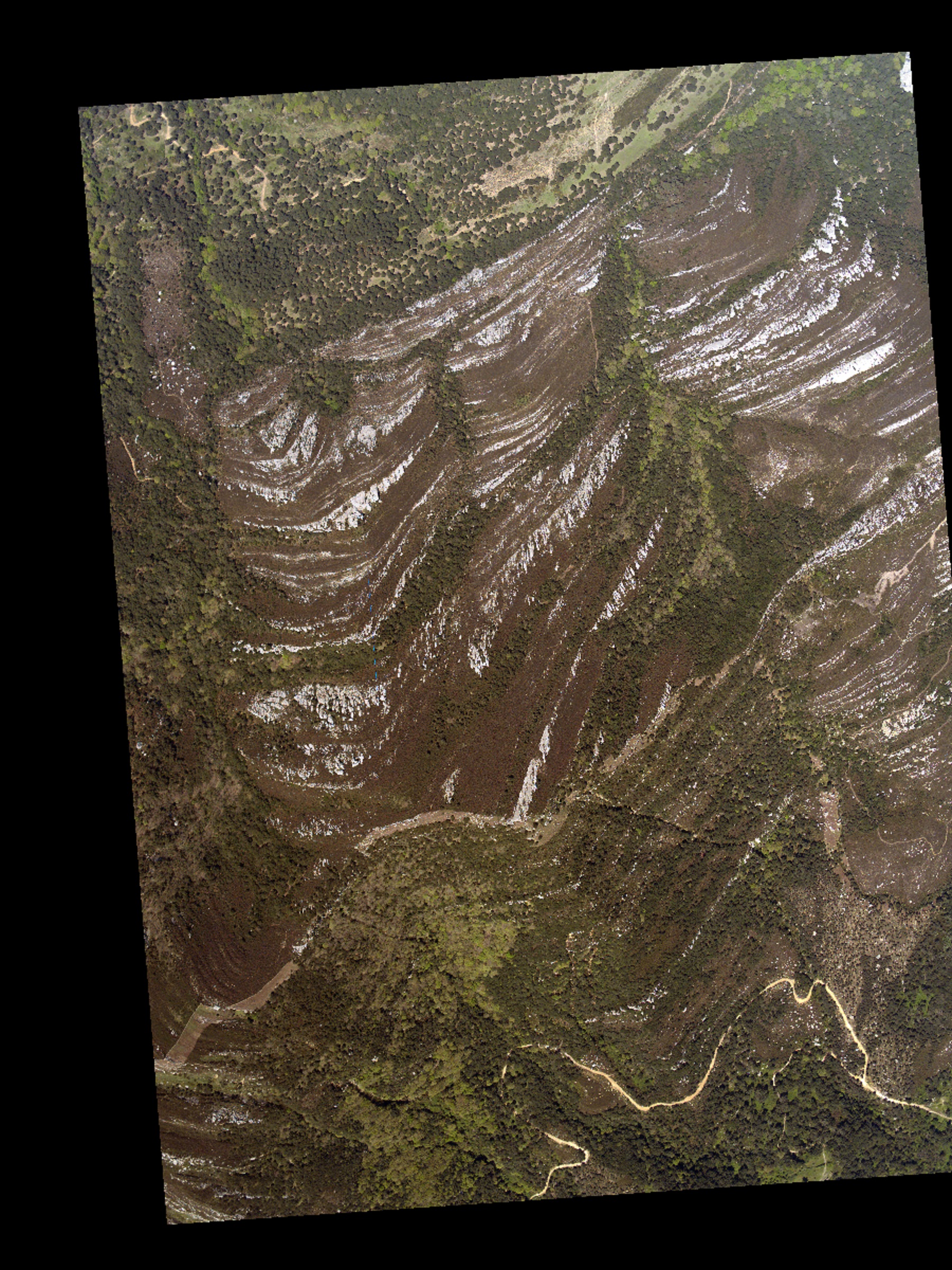} &
\includegraphics[width=40mm, height=55mm]{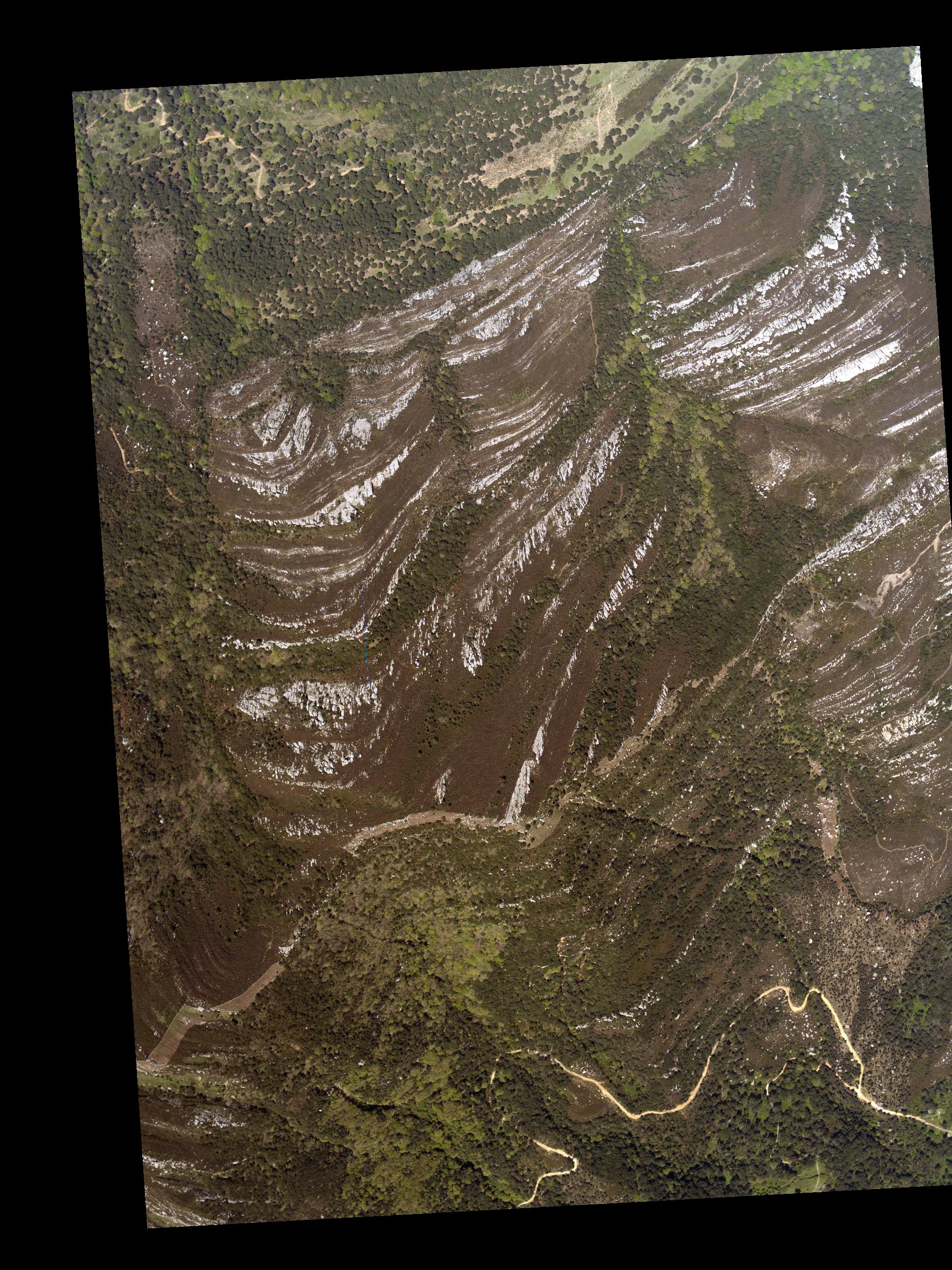} &
\includegraphics[width=40mm, height=55mm]{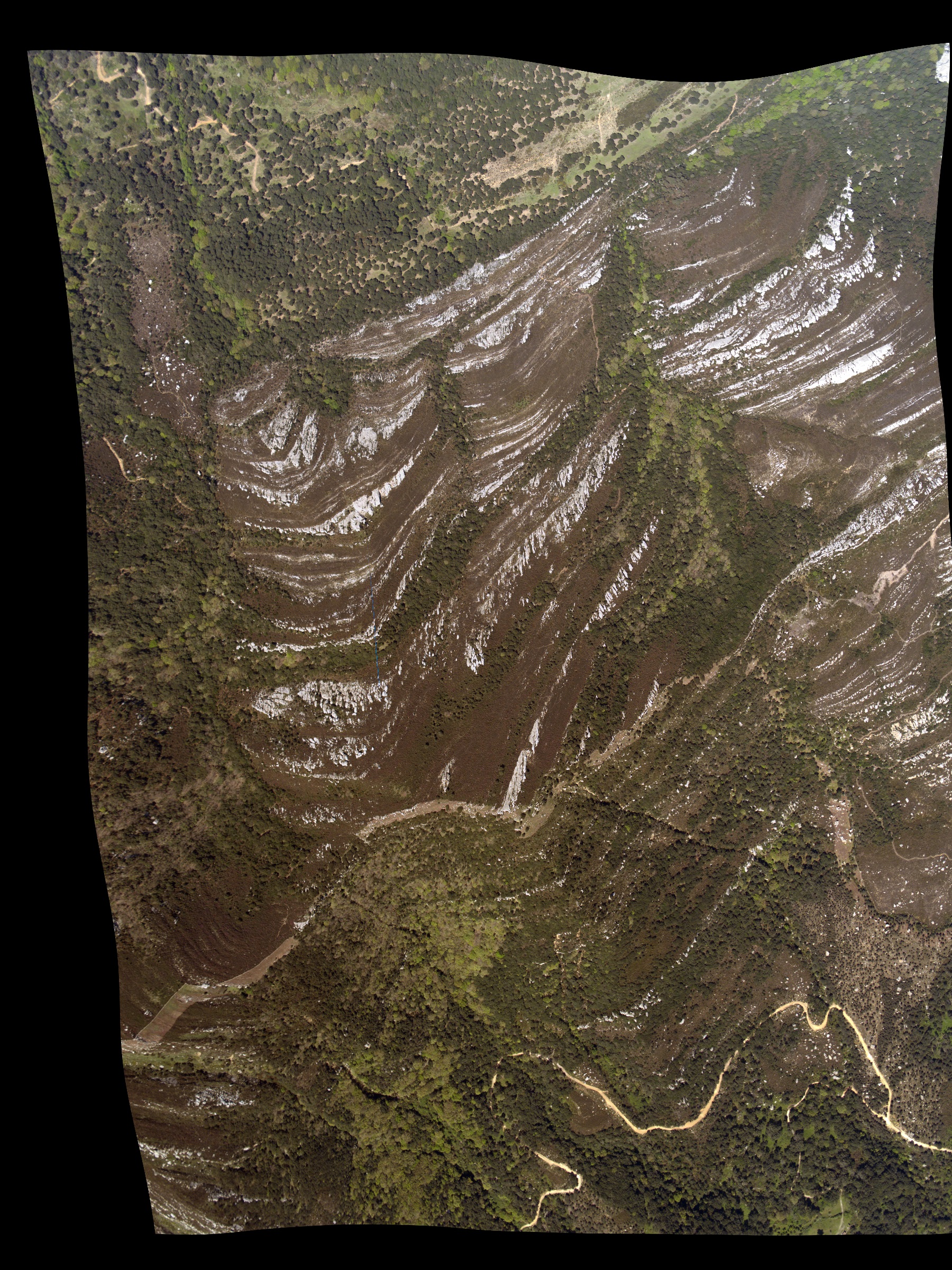} \\
 (d) NCC: $T^{\rm NCC}_{reg}$ & (e) MI: $T^{\rm MI}_{reg}$ &  
 (f) NGF: $T^{\rm NGF}_{reg}$ & (g) NP \eqref{eq:registfunc}: $T^{\rm NP}_{reg}$ \\
\includegraphics[width=40mm, height=55mm]{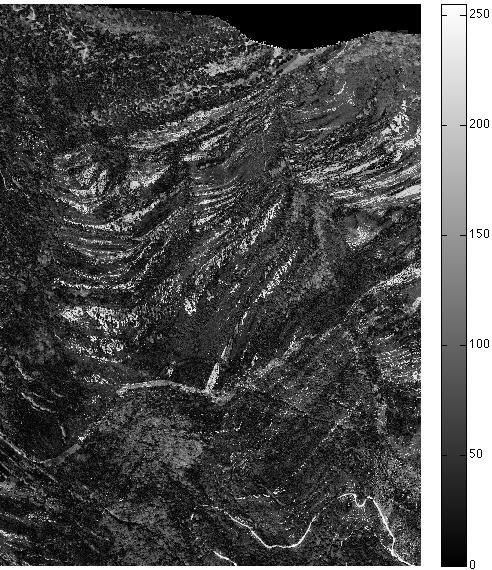} 
\put(-65,54){{\color{yellow} \circle{25,10}}} 
\put(-70,90){{\color{yellow} \circle{25,10}}} 
\put(-30,20){{\color{yellow} \circle{25,10}}}  &
\includegraphics[width=40mm, height=55mm]{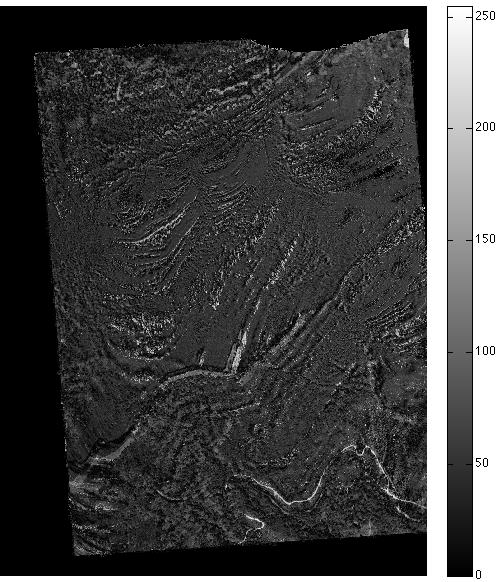} 
\put(-65,60){{\color{yellow} \circle{25,10}}} 
\put(-65,90){{\color{yellow} \circle{25,10}}} 
\put(-30,35){{\color{yellow} \circle{25,10}}}  &
\includegraphics[width=40mm, height=55mm]{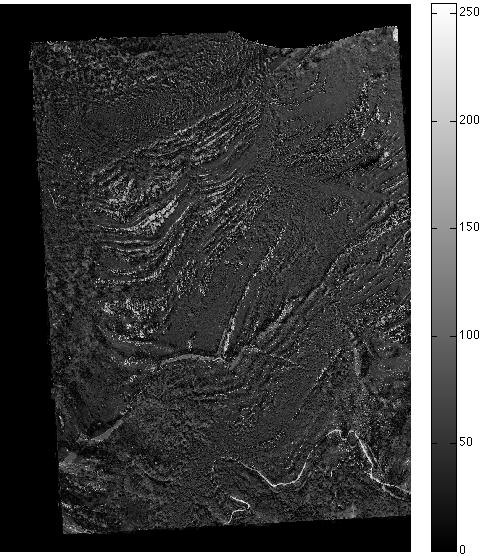} 
\put(-65,60){{\color{yellow} \circle{25,10}}} 
\put(-65,90){{\color{yellow} \circle{25,10}}} 
\put(-30,35){{\color{yellow} \circle{25,10}}}&
\includegraphics[width=40mm, height=55mm]{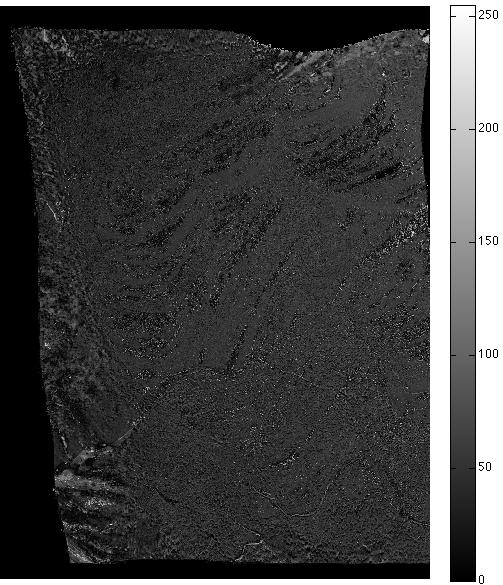} 
\put(-65,60){{\color{yellow} \circle{25,10}}} 
\put(-65,90){{\color{yellow} \circle{25,10}}} 
\put(-30,35){{\color{yellow} \circle{25,10}}} \\
 (h) $\vert T^{\rm NCC}_{reg}-R \vert$ (58.1) & (i)  $\vert T^{\rm MI}_{reg}-R \vert$ (49.9)
 & (j) $\vert T^{\rm NGF}_{reg}-R \vert$ (51.0) & (k) $\vert T^{\rm NP}_{reg}-R \vert$ (46.8)
\end{tabular}
\end{center}
\caption{Image registration of an aerial photograph onto a hyperspectral image 
in the case of rugged terrain (scale $2400 \times 1800$ m$^2$). The first row shows (a) 
a hyperspectral reference image ($R$); (b) an aerial photograph template image ($T$); 
(c) a map highlighting  difference between these images (i.e. the complement of intensity 
differences  $\vert T-R \vert$. The second row of panels show the aerial photograph 
template images after it has been registered using parametric methods (d) NCC, 
(e) MI and (f) NGF and (g) the NP approach. The final row shows maps highlighting the 
absolute values of the differences between the registered aerial photograph images and 
the hyperspectral reference image; the average intensity difference within the image is given in parentheses. 
}\label{example-rugged}
\end{figure*}

\begin{figure*}[t!]
\begin{center}
\begin{tabular}{cc}
\includegraphics[width=68mm, height=100mm]{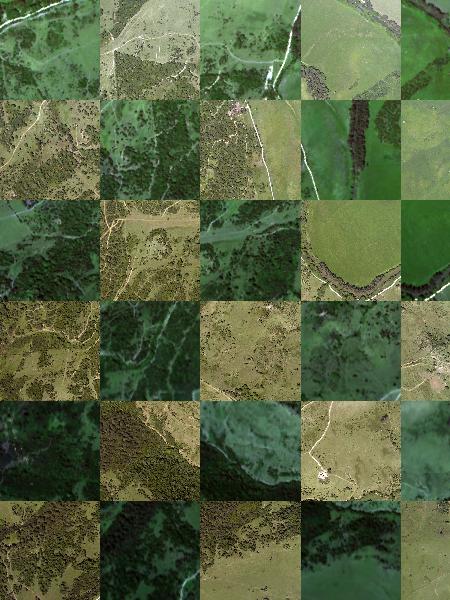}  &
\includegraphics[width=68mm, height=100mm]{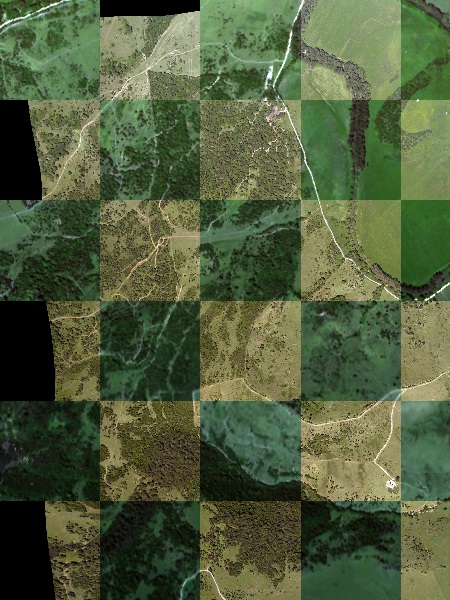} \\
(a) Raw image overlay in flat terrain case & 
(b) Registered image overlay in flat terrain case\\
\includegraphics[width=68mm, height=100mm]{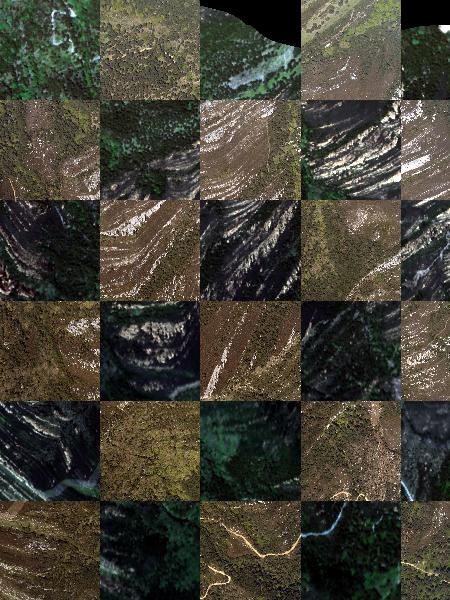}  &
\includegraphics[width=68mm, height=100mm]{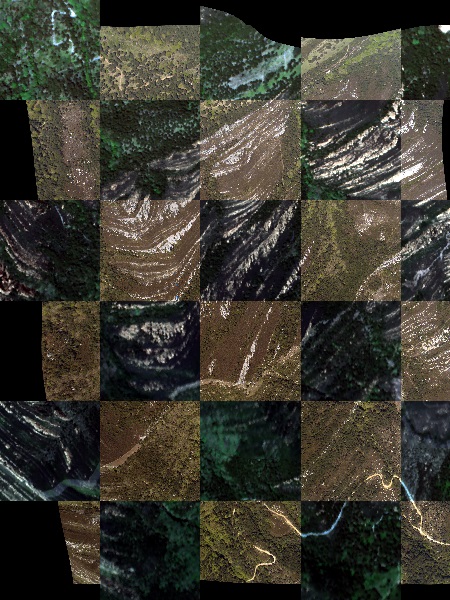}  \\
 (c) Raw image overlay in rugged terrain case & 
 (d) Registered image overlay in rugged terrain case 
\end{tabular}
\end{center}
\caption{Checkerboard overlay between aerial photograph template images ($T$) 
and hyperspectral reference images ($R$) and checkerboard overlay of registered 
aerial photograph template images ($T^{\rm NP}_{reg}$) and hyperspectral 
reference images ($R$) of the NP approach. (a)-(b) Flat terrain case; (c)-(d) rugged terrain case.
}\label{example-checkerboard}
\end{figure*}

Image registration of aerial photographs onto hyperspectral  or LiDAR images was more 
challenging  because the aerial photos were not preprocessed and did not come with 
georeferencing information. Moreover, the swath width of the aerial camera is much 
larger than that of LiDAR sensors, making it difficult to create a reference image 
onto which aerial photographs could be aligned. We present two image registration 
examples: one for a flat terrain and one for a rugged terrain (Figures \ref{example-flat} 
and \ref{example-rugged}, respectively). Where topographical variation 
is large the correct alignment of the images becomes more difficult 
\cite{bunting2008area,brunner2010earthquake,liang2014automatic,ye2014local}.  
.The non-parametric registration 
approach \eqref{eq:registfunc} worked well in the case of flat terrain  (see Figure \ref{example-flat}), 
while parametric registration
with three different distance measures (NCC, MI and NGF) poorly matched the detailed 
structures of a given reference image, see Figure \ref{example-flat} (h)--(k) in particular 
the parts marked by circles.
Approach \eqref{eq:registfunc} provides reasonable outcomes while 
parametric registration methods (NCC, MI and NGF) make serious mistakes and 
in particular, could not align detailed features (e.g. see red circles on 
Figure \ref{example-rugged} (h)--(k)). Figure \eqref{example-checkerboard} shows the 
results of aligning the aerial photographs onto the hyperspectral image for the 
cases of flat and rugged terrains in the form of a checkerboard:  if the aligment 
is good then features such as roads and rivers should join across the checkerboard.  
We can clearly see that the approach \eqref{eq:registfunc} gives very accurate registration results.

\begin{figure*}[t!]
\begin{center}
\begin{tabular}{c}
\includegraphics[width=175mm, height=38mm]{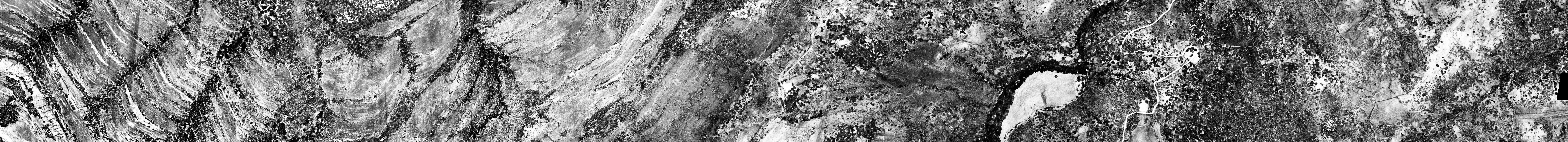}
\put(-135,70){{\color{red} \circle{25,10}}} \\
(a) LiDAR intensity image\\
\includegraphics[width=175mm, height=38mm]{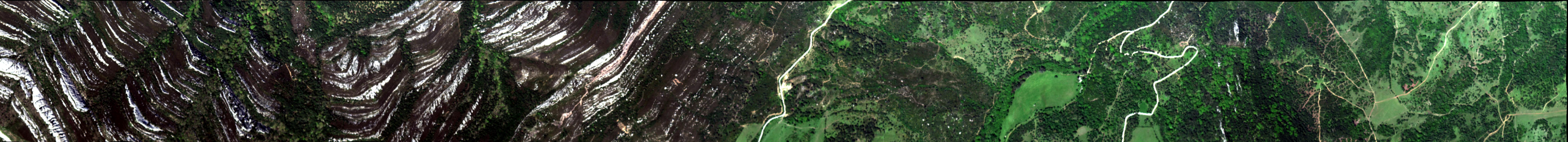}
\put(-135,70){{\color{red} \circle{25,10}}}\\
(b) RGB bands of registered hyperspectral image\\
\includegraphics[width=175mm, height=38mm]{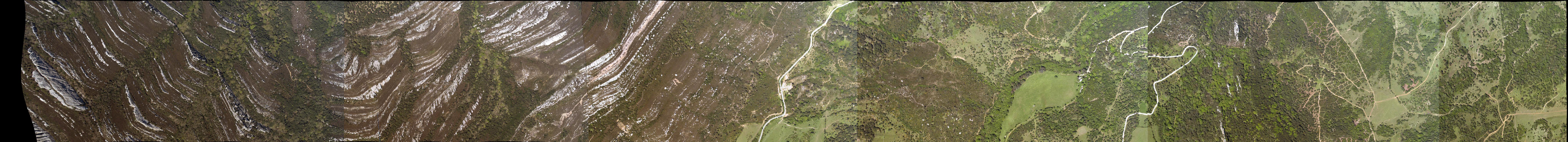} 
\put(-135,70){{\color{red} \circle{25,10}}}\\
(c) Mosaic image of registered aerial photographs (with fixed global parameter $\alpha$ in \eqref{eq:registfunc})\\
\includegraphics[width=175mm, height=38mm]{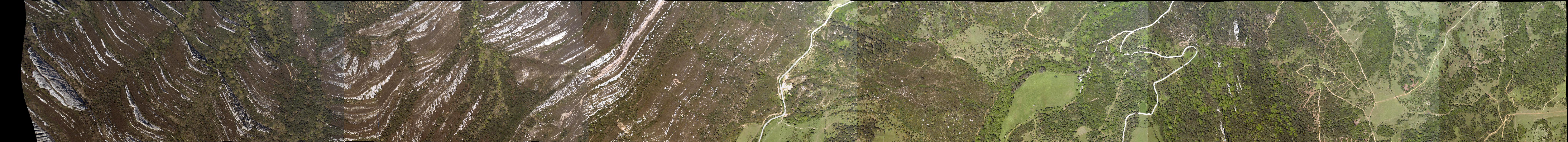} 
\put(-135,70){{\color{red} \circle{25,10}}}\\
(d) Mosaic image of registered aerial photographs (with locally tuned parameter $\alpha$ in \eqref{eq:registfunc})
\end{tabular}
\end{center}
\caption{Fully registered  LiDAR, hyperspectral and aerial photograph imagery. 
(a) LiDAR intensity image; (b) RGB bands of hyperspectral imagery; 
(c) mosaic imagery of registered aerial photographs of the NP approach with fixed global regularisation parameter $\alpha$;
(d) mosaic imagery of registered aerial photographs of the NP approach with locally tuned regularisation parameter $\alpha$.
}\label{example5}
\end{figure*}

After registration of aerial photographs onto hyperspectral images, additional registration 
was performed to remove minor mismatches between aerial photos and LiDAR (Figure \ref{example5}). 
As aerial photographs were registered individually onto hyperspectral imagery there may 
be mismatches at the edge of each aerial photograph (visible in  Figures \ref{example-flat} 
and \ref{example-rugged}), which  may produce 
noticeable discontinuity between the photographs.  For example, in  Figure \ref{example5} (c), 
the part marked by the red circle shows discontinuity at the interface of two aerial photographs. 
These boundary artefacts are due to a non-optimal choice of the regularisation parameter for 
the registration of aerial photographs to hyperspectral images.  We chose to have a fixed  
regularisation parameter $\alpha$ in \eqref{eq:registfunc} which might not be optimal for 
every aerial photo in the data set, and this  caused errors at the boundaries. 
Tuning the parameters for each aerial photograph where discontinuity deteriorates the 
quality of registration, can improve the 
result significantly. In the case of the mismatch inside the circle in Figure \ref{example5} (c) 
a tuning of the regularisation parameter $\alpha$ from $1.5\times 10^{5}$ to $2 \times 10^{5}$ 
significantly improved the registration and removed the discontinuity between 
the two aerial photos (Figure \ref{example5} (d)).

\section{Conclusions and Outlook}\label{sec:con} 
The experiments illustrated in  Figures \ref{example-flat}, \ref{example-rugged} 
and \ref{example-checkerboard} indicate that  non-parametric image registration techniques  
can effectively co-align remote sensing images well,  working as well as established methods 
when registration is straight forward and out-performing those approaches when dealing 
with non-georeferenced photos.   Remote sensing images are 
usually preprocessed before being sent to users, but the orthorectification and georeferencing 
procedures are not accurate enough to identify individual trees. 
Techniques based on feature extractions are well established in the field and 
are capable of accurate data assimilation if used in sufficient numbers, but this 
approach is at most semi-automatic. Intensity-based parametric methods, such as 
NCC, MI and NGF, can perform full-automatic registration but presumes that data 
are pre-processed and displacement between template and reference images is small. 
Non-parametric image registration provides flexible algorithms for image registration, 
thus allowing image registration without any prior knowledge of the dataset. 
The validation of this method in reducing processing time and improving 
analysis results was demonstrated by various experiments in multimodal remote sensing 
datasets, i.e., the LiDAR, hyperspectral and aerial photograph datasets.  

From the experiments shown in Section \ref{sec:experiments}, we see that the 
non-parametric registration with a variational formulation 
has successfully aligned remote sensing images. 
As it can be applied to non-orthorectified images, it performs automatic 
orthorectification and georeferencing as well. The non-parametric registration is of course
dependent on the quality of the reference image. We used LiDAR intensity 
images in order to register hyperspectral images onto LiDAR. We believe the quality 
of LiDAR can be further improved by increasing the understanding of the radiometric 
properties of LiDAR intensity. The automatic gain control system adjusts the pulse 
energy during the LiDAR acquisition (i.e. the pulse energy is increased when the 
returned energy is low). 
An AGC value within the range $[0, 255]$ is given for each pulse in the LAS file, 
and a few studies have attempted to normalise LiDAR intensity using these numbers 
\cite{kaasalainen2009radiometric, korpela2010range, korpela2010tree}. 
However, whilst none of those methods are able to successfully correct the LiDAR datasets we 
used,  we believe that a successful radiometric calibration could indeed improve the 
registration accuracy. Another difficulty is that hyperspectral and aerial photos are strongly 
influenced by shading effects, because they record backscattered solar energy. 
Shaded pixels create strong gradients on one side of trees, so the registration 
process is intrinsically biased to some extent. Hence, combining image registration with 
shade removal \cite{finlayson2009entropy} could improve the quality of image registration.

Similarity measures, such as Sum of Squared Distance (SSD), 
NCC or MI, could play a key role 
in image registration. SSD measures intensity difference between images and 
NCC maximises correlation between two images \cite{modersitzki2009fair}. 
Although these conventional methods can deal with images from the same 
measurement, their performance on multimodal images are poor. 
The MI method has been widely used as a similarity measure in remote 
sensing applications as it can be applied to multimodal imaging. 
MI, which is originally from information theory, measures joint probability 
of intensities of images, can be viewed as a generalised similarity measure 
\cite{modersitzki2009fair}. However, the MI method has noticeable disadvantages. 
MI is highly non-convex, therefore it is difficult to optimise and increases non-linearity 
of registration \cite{haber2006intensity}. Since MI is based on joint density of 
intensity values, it may suffer from interpolation induced artefact 
\cite{chen2003performance}. The NGF method is designed to 
measure similarity between images taken by multi-sensors. It compares gradients 
of two images, so it is computationally fast and handles multimodality. 

Regularisation is a key part of this paper. Although a number of studies have used 
intensity-based similarity measures 
 \cite{zitova2003image,chen2003mutual,chen2003performance,suri2010mutual,
 parmehr2014automatic}, the ill-posedness of these measures prevents their use in flexible applications 
in remote sensing. This means that successful image 
registration is conditional upon the data being preprocessed and displacement between images being small. 
In theory, adding a regularisation term in addition to similarity term makes the problem 
close to or exactly well-posed such that the registration problem has 
a much more meaningful solution, although 
it is difficult to remove all local minima to get the exact solution in reality. 
A few regularisation methods have been suggested to guarantee well-posedness during 
the registration process \cite{burger2013hyperelastic}. As we mentioned before, 
many regularisation techniques are sensitive to affine linear displacement such 
that pre-registration with affine linear transformation is required 
\cite{modersitzki2003numerical,fischer2003curvature,fischer2004unified}. 
In contrast, this current  research uses curvature regularisation, which does not require affine 
preregistration steps. However, pre-registration at coarsest level is recommended in 
general applications as non-parametric registration still penalises affine transformation 
by its boundary conditions (i.e. it is still influenced by initial 
position of two images to some extent, see\cite{henn2006full,modersitzki2009fair}).

\noindent {\bf Acknowledgements:}
The authors would like to thank NERC-ARSF for collecting and pre-processing the data used in this  
research project [EU11/03/100], and the grants supported from King Abdullah 
University of Science Technology and Wellcome Trust. We thank Ben Taylor of Plymouth Marine Laboratoy and Will Simonson for their valuable comments on the manuscript.



\bibliographystyle{plain}
\bibliography{bibliography_NPIR}

\begin{thebibliography}{10}

\bibitem{amit1994nonlinear}
Y.~Amit.
\newblock A nonlinear variational problem for image matching.
\newblock {\em SIAM Journal on Scientific Computing}, 15(1):207--224, 1994.

\bibitem{ascher2006effective}
U.~Ascher, E.~Haber, and H.~Huang.
\newblock On effective methods for implicit piecewise smooth surface recovery.
\newblock {\em SIAM Journal on Scientific Computing}, 28(1):339--358, 2006.

\bibitem{asner2009tropical}
G.~Asner.
\newblock Tropical forest carbon assessment: integrating satellite and airborne
  mapping approaches.
\newblock {\em Environmental Research Letters}, 4(3):034009, 2009.

\bibitem{asner2007carnegie}
G.~Asner, J.~Boardman, C.~Field, D.~Knapp, T.~Kennedy-Bowdoin, M.~Jones, and
  R.~Martin.
\newblock Carnegie airborne observatory: in-flight fusion of hyperspectral
  imaging and waveform light detection and ranging for three-dimensional
  studies of ecosystems.
\newblock {\em Journal of Applied Remote Sensing}, 1(1):013536--013536, 2007.

\bibitem{asner2008pnas}
G.~Asner, R.~Hughes, P.~Vitousek, D.~Knapp, T.~Kennedy-Bowdoin, J.~Boardman,
  R.~Martin, M.~Eastwood, and R.~Green.
\newblock Invasive plants transform the three-dimensional structure of rain
  forests.
\newblock {\em Proceedings of the National Academy of Sciences},
  105(11):4519--4523, 2008.

\bibitem{asner2008invasive}
G.~Asner, D.~Knapp, T.~Kennedy-Bowdoin, M.~Jones, R.~Martin, J.~Boardman, and
  R.~Hughes.
\newblock Invasive species detection in hawaiian rainforests using airborne
  imaging spectroscopy and lidar.
\newblock {\em Remote Sensing of Environment}, 112(5):1942--1955, 2008.

\bibitem{asner2011canopy}
G.~Asner and R.~Martin.
\newblock Canopy phylogenetic, chemical and spectral assembly in a lowland
  amazonian forest.
\newblock {\em New Phytologist}, 189(4):999--1012, 2011.

\bibitem{bentoutou2005automatic}
Y.~Bentoutou, N.~Taleb, K.~Kpalma, and J.~Ronsin.
\newblock An automatic image registration for applications in remote sensing.
\newblock {\em Geoscience and Remote Sensing, IEEE Transactions on},
  43(9):2127--2137, 2005.

\bibitem{berni2009thermal}
J.~Berni, P.~Zarco-Tejada, L.~Su{\'a}rez, and E.~Fereres.
\newblock Thermal and narrowband multispectral remote sensing for vegetation
  monitoring from an unmanned aerial vehicle.
\newblock {\em Geoscience and Remote Sensing, IEEE Transactions on},
  47(3):722--738, 2009.

\bibitem{bro1996fast}
M.~Bro-Nielsen and C.~Gramkow.
\newblock Fast fluid registration of medical images.
\newblock In {\em Visualization in Biomedical Computing}, pages 265--276.
  Springer, 1996.

\bibitem{broit1981optimal}
C.~Broit.
\newblock {\em Optimal registration of deformed images}.
\newblock PhD thesis, 1981.

\bibitem{brown1992survey}
L.~Brown.
\newblock A survey of image registration techniques.
\newblock {\em ACM computing surveys (CSUR)}, 24(4):325--376, 1992.

\bibitem{brunner2010earthquake}
D.~Brunner, G.~Lemoine, and L.~Bruzzone.
\newblock Earthquake damage assessment of buildings using vhr optical and sar
  imagery.
\newblock {\em Geoscience and Remote Sensing, IEEE Transactions on},
  48(5):2403--2420, 2010.

\bibitem{bryson2010airborne}
M.~Bryson, A.~Reid, F.~Ramos, and S.~Sukkarieh.
\newblock Airborne vision-based mapping and classification of large farmland
  environments.
\newblock {\em Journal of Field Robotics}, 27(5):632--655, 2010.

\bibitem{B87}
M.~Bueger.
\newblock {\em Geometry I}.
\newblock Berlin: Springer, 1987.

\bibitem{bunting2008area}
P.~Bunting, R.~Lucas, and F.~Labrosse.
\newblock An area based technique for image-to-image registration of
  multi-modal remote sensing data.
\newblock In {\em Geoscience and Remote Sensing Symposium, 2008. IGARSS 2008.
  IEEE International}, volume~5, pages V--212. IEEE, 2008.

\bibitem{burger2013hyperelastic}
M.~Burger, J.~Modersitzki, and L.~Ruthotto.
\newblock A hyperelastic regularization energy for image registration.
\newblock {\em SIAM Journal on Scientific Computing}, 35(1):B132--B148, 2013.

\bibitem{chen2003performance}
H.~Chen, P.~Varshney, and M.~Arora.
\newblock Performance of mutual information similarity measure for registration
  of multitemporal remote sensing images.
\newblock {\em Geoscience and Remote Sensing, IEEE Transactions on},
  41(11):2445--2454, 2003.

\bibitem{chen2003mutual}
H-M Chen, Manoj~K Arora, and Pramod~K Varshney.
\newblock Mutual information-based image registration for remote sensing data.
\newblock {\em International Journal of Remote Sensing}, 24(18):3701--3706,
  2003.

\bibitem{christensen1994deformable}
G.~Christensen.
\newblock {\em Deformable shape models for anatomy}.
\newblock PhD thesis, Washington University Saint Louis, Mississippi, 1994.

\bibitem{cole2003multiresolution}
A.~Cole-Rhodes, K.~Johnson, J.~Le Moigne, and I.~Zavorin.
\newblock Multiresolution registration of remote sensing imagery by
  optimization of mutual information using a stochastic gradient.
\newblock {\em Image Processing, IEEE Transactions on}, 12(12):1495--1511,
  2003.

\bibitem{dalponte2008fusion}
M.~Dalponte, L.~Bruzzone, and D.~Gianelle.
\newblock Fusion of hyperspectral and lidar remote sensing data for
  classification of complex forest areas.
\newblock {\em Geoscience and Remote Sensing, IEEE Transactions on},
  46(5):1416--1427, 2008.

\bibitem{finlayson2009entropy}
G.~Finlayson, M.~Drew, and C.~Lu.
\newblock Entropy minimization for shadow removal.
\newblock {\em International Journal of Computer Vision}, 85(1):35--57, 2009.

\bibitem{fischer2003curvature}
B.~Fischer and J.~Modersitzki.
\newblock Curvature based image registration.
\newblock {\em Journal of Mathematical Imaging and Vision}, 18(1):81--85, 2003.

\bibitem{fischer2004unified}
B.~Fischer and J.~Modersitzki.
\newblock A unified approach to fast image registration and a new curvature
  based registration technique.
\newblock {\em Linear Algebra and its applications}, 380:107--124, 2004.

\bibitem{fischler1981random}
M.~Fischler and R.~Bolles.
\newblock Random sample consensus: a paradigm for model fitting with
  applications to image analysis and automated cartography.
\newblock {\em Communications of the ACM}, 24(6):381--395, 1981.

\bibitem{fonseca1996registration}
L.~Fonseca and BS. Manjunath.
\newblock Registration techniques for multisensor remotely sensed imagery.
\newblock {\em PE \& RS- Photogrammetric Engineering \& Remote Sensing},
  62(9):1049--1056, 1996.

\bibitem{goncalves2011automatic}
H.~Goncalves, L.~Corte-Real, and J.~Goncalves.
\newblock Automatic image registration through image segmentation and sift.
\newblock {\em Geoscience and Remote Sensing, IEEE Transactions on},
  49(7):2589--2600, 2011.

\bibitem{gonccalves2011hairis}
H.~Gon{\c{c}}alves, J.~Gon{\c{c}}alves, and L.~Corte-Real.
\newblock Hairis: A method for automatic image registration through
  histogram-based image segmentation.
\newblock {\em Image Processing, IEEE Transactions on}, 20(3):776--789, 2011.

\bibitem{haber2006intensity}
E.~Haber and J.~Modersitzki.
\newblock Intensity gradient based registration and fusion of multi-modal
  images.
\newblock In {\em Medical Image Computing and Computer-Assisted
  Intervention--MICCAI 2006}, pages 726--733. Springer, 2006.

\bibitem{haber2006multilevel}
E.~Haber and J.~Modersitzki.
\newblock A multilevel method for image registration.
\newblock {\em SIAM Journal on Scientific Computing}, 27(5):1594--1607, 2006.

\bibitem{hadamard1902problemes}
J.~Hadamard.
\newblock Sur les probl{\`e}mes aux d{\'e}riv{\'e}es partielles et leur
  signification physique.
\newblock {\em Princeton university bulletin}, 13(49-52):28, 1902.

\bibitem{henn2006full}
S.~Henn.
\newblock A full curvature based algorithm for image registration.
\newblock {\em Journal of Mathematical Imaging and Vision}, 24(2):195--208,
  2006.

\bibitem{hong2008wavelet}
G.~Hong and Y.~Zhang.
\newblock Wavelet-based image registration technique for high-resolution remote
  sensing images.
\newblock {\em Computers \& Geosciences}, 34(12):1708--1720, 2008.

\bibitem{huang2009estimating}
S.~Huang, R.~Crabtree, C.~Potter, and P.~Gross.
\newblock Estimating the quantity and quality of coarse woody debris in
  yellowstone post-fire forest ecosystem from fusion of sar and optical data.
\newblock {\em Remote Sensing of Environment}, 113(9):1926--1938, 2009.

\bibitem{hudak1998textural}
A.~Hudak and C.~Wessman.
\newblock Textural analysis of historical aerial photography to characterize
  woody plant encroachment in south african savanna.
\newblock {\em Remote Sensing of Environment}, 66(3):317--330, 1998.

\bibitem{inglada2004possibility}
J.~Inglada and A.~Giros.
\newblock On the possibility of automatic multisensor image registration.
\newblock {\em Geoscience and Remote Sensing, IEEE Transactions on},
  42(10):2104--2120, 2004.

\bibitem{kaasalainen2009radiometric}
S.~Kaasalainen, H.~Hyyppa, A.~Kukko, P.~Litkey, E.~Ahokas, J.~Hyyppa,
  H.~Lehner, A.~Jaakkola, J.~Suomalainen, A.~Akujarvi, M.~Kaasalainen, and
  U.~Pyysalo.
\newblock Radiometric calibration of lidar intensity with commercially
  available reference targets.
\newblock {\em Geoscience and Remote Sensing, IEEE Transactions on},
  47(2):588--598, 2009.

\bibitem{kim2003automatic}
T.~Kim and Y.~Im.
\newblock Automatic satellite image registration by combination of matching and
  random sample consensus.
\newblock {\em Geoscience and Remote Sensing, IEEE Transactions on},
  41(5):1111--1117, 2003.

\bibitem{korpela2010range}
I.~Korpela, H.~{\O}rka, J.~Hyypp{\"a}, V.~Heikkinen, and T.~Tokola.
\newblock Range and agc normalization in airborne discrete-return lidar
  intensity data for forest canopies.
\newblock {\em ISPRS Journal of Photogrammetry and Remote Sensing},
  65(4):369--379, 2010.

\bibitem{korpela2010tree}
I.~Korpela, H.~{\O}rka, M.~Maltamo, T.~Tokola, J.~Hyypp{\"a}, and etal.
\newblock Tree species classification using airborne lidar--effects of stand
  and tree parameters, downsizing of training set, intensity normalization, and
  sensor type.
\newblock {\em Silva Fennica}, 44(2):319--339, 2010.

\bibitem{kroon2009mri}
D.~Kroon and C.~Slump.
\newblock Mri modalitiy transformation in demon registration.
\newblock In {\em Biomedical Imaging: From Nano to Macro, 2009. ISBI'09. IEEE
  International Symposium on}, pages 963--966. IEEE, 2009.

\bibitem{laliberte2010acquisition}
A.~Laliberte, J.~Herrick, A.~Rango, and C.~Winters.
\newblock Acquisition, orthorectification, and object-based classification of
  unmanned aerial vehicle (uav) imagery for rangeland monitoring.
\newblock {\em Photogrammetric Engineering and Remote Sensing}, 76(6):661--672,
  2010.

\bibitem{lefsky2002lidar}
M.~Lefsky, W.~Cohen, G.~Parker, and D.~Harding.
\newblock Lidar remote sensing for ecosystem studies lidar, an emerging remote
  sensing technology that directly measures the three-dimensional distribution
  of plant canopies, can accurately estimate vegetation structural attributes
  and should be of particular interest to forest, landscape, and global
  ecologists.
\newblock {\em BioScience}, 52(1):19--30, 2002.

\bibitem{li1995contour}
H.~Li, B.~Manjunath, and S.~Mitra.
\newblock A contour-based approach to multisensor image registration.
\newblock {\em Image Processing, IEEE Transactions on}, 4(3):320--334, 1995.

\bibitem{li2009robust}
Q.~Li, G.~Wang, J.~Liu, and S.~Chen.
\newblock Robust scale-invariant feature matching for remote sensing image
  registration.
\newblock {\em Geoscience and Remote Sensing Letters, IEEE}, 6(2):287--291,
  2009.

\bibitem{liang2014automatic}
J.~Liang, X.~Liu, K.~Huang, X.~Li, D.~Wang, and X.~Wang.
\newblock Automatic registration of multisensor images using an integrated
  spatial and mutual information (smi) metric.
\newblock {\em Geoscience and Remote Sensing, IEEE Transactions on},
  51(1):603--615, 2014.

\bibitem{lim2003lidar}
K.~Lim, P.~Treitz, M.~Wulder, B.~St-Onge, and M.~Flood.
\newblock Lidar remote sensing of forest structure.
\newblock {\em Progress in Physical Geography}, 27(1):88--106, 2003.

\bibitem{lin2007map}
Y.~Lin and G.~Medioni.
\newblock Map-enhanced uav image sequence registration and synchronization of
  multiple image sequences.
\newblock In {\em Computer Vision and Pattern Recognition, 2007. CVPR'07. IEEE
  Conference on}, pages 1--7. IEEE, 2007.

\bibitem{lowe2004distinctive}
D.~Lowe.
\newblock Distinctive image features from scale-invariant keypoints.
\newblock {\em International journal of computer vision}, 60(2):91--110, 2004.

\bibitem{maintz1998survey}
J.~Maintz and M.~Viergever.
\newblock A survey of medical image registration.
\newblock {\em Medical image analysis}, 2(1):1--36, 1998.

\bibitem{modersitzki2003numerical}
J.~Modersitzki.
\newblock {\em Numerical methods for image registration}.
\newblock OUP Oxford, 2003.

\bibitem{modersitzki2009fair}
J.~Modersitzki.
\newblock {\em FAIR: flexible algorithms for image registration}, volume~6.
\newblock SIAM, 2009.

\bibitem{le2011image}
J.~Le Moigne, N.~Netanyahu, and R.~Eastman.
\newblock {\em Image registration for remote sensing}.
\newblock Cambridge University Press, 2011.

\bibitem{nakashizuka1995forest}
T.~Nakashizuka, T.~Katsuki, and H.~Tanaka.
\newblock Forest canopy structure analyzed by using aerial photographs.
\newblock {\em Ecological Research}, 10(1):13--18, 1995.

\bibitem{parmehr2014automatic}
E.~Parmehr, C.~Fraser, C.~Zhang, and J.~Leach.
\newblock Automatic registration of optical imagery with 3d lidar data using
  statistical similarity.
\newblock {\em ISPRS Journal of Photogrammetry and Remote Sensing}, 88:28--40,
  2014.

\bibitem{song2014novel}
Zhili Song, Shuigeng Zhou, and Jihong Guan.
\newblock A novel image registration algorithm for remote sensing under affine
  transformation.
\newblock 2014.

\bibitem{suri2010mutual}
S.~Suri and P.~Reinartz.
\newblock Mutual-information-based registration of terrasar-x and ikonos
  imagery in urban areas.
\newblock {\em Geoscience and Remote Sensing, IEEE Transactions on},
  48(2):939--949, 2010.

\bibitem{turner2014direct}
D.~Turner, A.~Lucieer, and L.~Wallace.
\newblock Direct georeferencing of ultrahigh-resolution uav imagery.
\newblock 52(5):2738--2745, 2014.

\bibitem{turner2012automated}
D.~Turner, A.~Lucieer, and C.~Watson.
\newblock An automated technique for generating georectified mosaics from
  ultra-high resolution unmanned aerial vehicle (uav) imagery, based on
  structure from motion (sfm) point clouds.
\newblock {\em Remote Sensing}, 4(5):1392--1410, 2012.

\bibitem{verhoeven2012mapping}
G.~Verhoeven, M.~Doneus, C.~Briese, and F.~Vermeulen.
\newblock Mapping by matching: a computer vision-based approach to fast and
  accurate georeferencing of archaeological aerial photographs.
\newblock {\em Journal of Archaeological Science}, 39(7):2060--2070, 2012.

\bibitem{wahed2013automatic}
M.~Wahed, G.~El-tawel, and A.~El-karim.
\newblock Automatic image registration technique of remote sensing images.
\newblock {\em International Journal of Advanced Computer Science \&
  Applications}, 4(2), 2013.

\bibitem{wong2007arrsi}
A.~Wong and D.~Clausi.
\newblock Arrsi: automatic registration of remote-sensing images.
\newblock {\em Geoscience and Remote Sensing, IEEE Transactions on},
  45(5):1483--1493, 2007.

\bibitem{yang2009remote}
Y.~Yang and X.~Gao.
\newblock Remote sensing image registration via active contour model.
\newblock {\em AEU-international journal of electronics and communications},
  63(4):227--234, 2009.

\bibitem{ye2014local}
Y.~Ye and J.~Shan.
\newblock A local descriptor based registration method for multispectral remote
  sensing images with non-linear intensity differences.
\newblock {\em ISPRS Journal of Photogrammetry and Remote Sensing}, 90:83--95,
  2014.

\bibitem{zitova2003image}
B.~Zitova and J.~Flusser.
\newblock Image registration methods: a survey.
\newblock {\em Image and vision computing}, 21(11):977--1000, 2003.

\end{thebibliography}

\end{document}